\documentclass[11pt]{article}

\usepackage[preprint]{acl}

\usepackage{times}
\usepackage{latexsym}

\usepackage[T1]{fontenc}

\usepackage[utf8]{inputenc}
\usepackage{amsmath}
\usepackage{microtype}

\usepackage{inconsolata}

\usepackage{graphicx}
\usepackage{enumitem}
\usepackage{amsmath,amssymb,amsfonts}
\usepackage{booktabs}
\usepackage{adjustbox}
\usepackage{multirow}
\usepackage[table]{xcolor}
\usepackage[most]{tcolorbox}

\definecolor{asblue}{RGB}{220,235,255} 
\newtcolorbox{responsebox}[2][]{
  enhanced,
  breakable,
  colback=green!5,
  colframe=green!65!black,
  coltitle=white,
  fonttitle=\bfseries\small\ttfamily\color{white},
  fontupper=\ttfamily\footnotesize,
  boxrule=0.8pt,
  title={#2},
  title style={fill=green!65!black},
  left=5mm, right=5mm,
  top=3mm, bottom=3mm,
  arc=2mm,
  borderline west={3pt}{0pt}{green!75},
  boxsep=2pt,
  #1
}

\newtcolorbox{promptbox}[2][]{
  enhanced,
  breakable,
  colback=gray!5,
  colframe=gray!65!black,
  coltitle=white,
  fonttitle=\bfseries\small\ttfamily\color{white},
  fontupper=\ttfamily\footnotesize,
  boxrule=0.8pt,
  title={#2},
  title style={fill=gray!65!black},
  left=5mm, right=5mm,
  top=3mm, bottom=3mm,
  arc=2mm,
  borderline west={3pt}{0pt}{gray!75},
  boxsep=2pt,
  #1
}

\title{Agent Skill Framework: Perspectives on the Potential of Small to Medium Language Models in Industrial Environments}

\author{
 \textbf{Yangjie Xu\textsuperscript{$\spadesuit$1}},
 \textbf{Lujun Li\textsuperscript{$\spadesuit$1}},
 \textbf{Lama Sleem\textsuperscript{1}},
 \textbf{Niccolo' Gentile\textsuperscript{2}},
 \\
  \textbf{Yewei Song\textsuperscript{1}},
 \textbf{Yiqun Wang\textsuperscript{1}},
 \textbf{Siming Ji\textsuperscript{3}},
 \textbf{Wenbo Wu\textsuperscript{4}},
 \textbf{Hui Huang\textsuperscript{5}},
 \textbf{Radu State\textsuperscript{1}},
\\
 \textsuperscript{1}University of Luxembourg,
 \textsuperscript{2}Foyer S.A.,
 \textsuperscript{3}Princeton University,
 \textsuperscript{4}Université Paris-Saclay,
 \textsuperscript{5}LIST,
\\
}

\begin{document}

\maketitle

\begin{abstract}
Skills are getting increasingly used in recent agentic frameworks, performing robustly with proprietary LLMs. Yet, their effectiveness for small and medium-sized open-source models (270M--80B) remains underexplored, especially under industrial, resource-constrained conditions. As such, we present a large-scale empirical study evaluating the skill paradigm's feasibility across two open-source classification tasks: one insurance-claim task based on proprietary data, and a set of Python programming problems. We find that very small models frequently fail to retrieve or select correct skills, substantially limiting their usefulness in the evaluated skill-augmented settings, while larger 30B--80B models gain substantial benefits from skill augmentation. Furthermore, we provide a deployment-oriented GPU profile, reporting VRAM time occupancy (GB$\cdot$min) across models under our evaluation settings---a deployment-oriented metric that may help characterize GPU-memory costs, where small and medium models are typically preferred for VRAM control. These results characterize how the benefits of skill augmentation vary with model scale, task complexity, and GPU budget, and provide deployment-oriented observations.


\end{abstract}

\section{Introduction}

Agent skills (commonly stored as \texttt{SKILL.md} files in agentic systems), initially introduced by Anthropic, are increasingly adopted in practice for coding agents and personal AI assistants. This approach can be viewed as a well-designed \emph{static cheat sheet} \cite{suzgun2025_DynamicCheatsheet}: it provides large language models (LLMs) with reusable, domain-specific procedural knowledge, directs their attention toward salient task-relevant information and guidelines, and facilitates progressive context management for long-horizon task execution. This approach substantially optimizes both the length and quality of context windows while improving LLMs' use of tools and external knowledge resources. Unlike Retrieval Augmented Generation (RAG) systems \citep{NEURIPS2020_6b493230}, which rigidly encode textual or visual data into vector-space databases, the skills design directly leverages the In-Context Learning (ICL) capabilities of LLMs \citep{dong-etal-2024-survey,DBLP:journals/corr/abs-2005-14165} to dynamically select the most pertinent information, instructions, and contextual knowledge.

\begin{figure}[htbp]
    \centering
    \includegraphics[width=0.45\textwidth]{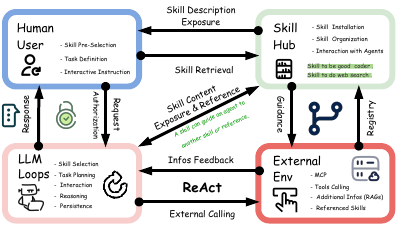}
    \caption{Agent framework with skill}
    \label{fig:skill-agent-framework}
\end{figure}

However, the capacity to leverage reasoning capabilities depends heavily on the inherent performance of large-size foundation models such as GPT or Claude, and existing approaches are primarily tailored to closed-source models specifically optimized for code-related tasks \cite{zhang2024codeagentenhancingcodegeneration}. In addition, concerns have been raised regarding data security \cite{YAN2025100300}, as well as the \textbf{substantial computational and financial costs} associated with invoking proprietary APIs or locally serving large-size models. Moreover, it remains largely unclear whether agent skills can deliver comparable benefits when deployed on open-source, limited-size models—particularly small-to-medium language models. Consequently, our contributions are:
\begin{itemize}
    \item A large-scale empirical evaluation of agent skills across open-source models from 270M to 80B parameters, on public benchmarks and \textsc{InsurBench}, a proprietary decision-making benchmark based on real insurance email threads. 
    \item A deployment-oriented cost analysis that jointly considers task accuracy, latency, and GPU memory--time (GB·min), providing practical guidance under constrained budgets.
    \item Exploratory studies on practical design choices for skill-augmented agents, including (i) scaling the skill hub from a few to over 100 skills, (ii) incorporating chat history into agent decisions, and (iii) replacing the keyword ``skill'' with alternative terms such as ``capability'' and ``expertise''.
\end{itemize}

\section{Related Work}

\subsection{Context Engineering (CE)}
The concept of CE has already been widely researched, and multiple engineering practices across multiple directions have been suggested. With the advent of zero-shot or few-shot generalization capabilities in LLMs \cite{brown2020languagemodelsfewshotlearners, dong2024surveyincontextlearning}, post-training paradigms have gradually been challenged by more convenient, efficient, and cost-effective post-deployment CE approaches for domain adaptation and agent behavior improvement. Moreover, a growing body of research has validated that LLMs exhibit human-like attention limitations and the "Lost in the Middle" / "Context rot" \cite{anthropic2025context} phenomenon when confronted with excessively long contexts \cite{Kou_2024,dai2024deniahlincontextfeaturesinfluence,du2025contextlengthhurtsllm}, demonstrating the importance of curated CE practices. Contemporary LLMs as such employ various CE designs, including but not limited to hierarchical multi-agent systems to handle complex tasks\cite{luo2025largelanguagemodelagent}, routing steps directly fed to designated agents \cite{yue-etal-2025-masrouter}, and sophisticated management approaches to multi-turn dialogue histories. These CE techniques are integrated within heterogeneously long and short term agent memory modules \cite{zhang2024surveymemorymechanismlarge,salama-etal-2025-meminsight,hu-etal-2025-hiagent} leveraging file systems, vector databases, knowledge graphs \cite{edge2025localglobalgraphrag}, and memory buffers \cite{xu2025amemagenticmemoryllm} to achieve comprehensive enhancements in contextual coherence, personalized learning, and complex task decision-making.

\subsection{Agentic systems with skills}

\citeauthor{ye2026metacontextengineeringagentic} conceptualizes CE as an evolving skill, proposing a two-layer framework termed "Meta CE" to automatically rewrite and optimize skill descriptions. \citeauthor{li2026singleagentskillsreplacemultiagent} demonstrates that, for many reasoning tasks, constructing a single-agent system equipped with a skill library achieves accuracy comparable to multi-agent systems while reducing token consumption and latency by approximately half. \cite{li2026skillsbenchbenchmarkingagentskills} introduces SkillsBench to systematically evaluate whether endowing agents with skills is truly worthwhile, but it focuses primarily on closed-source large models.
Numerous open-source libraries for agent skills, such as DeepAgents and the agent skill Collections\footnote{https://github.com/heilcheng/awesome-agent-skills}, have grown rapidly in popularity \cite{chen2026cuaskilldevelopskillscomputer}, while also sparking discussions on skill safety and permissions \citeauthor{liu2026agentskillswildempirical}. When using agent skills, model selection proves crucial for skill routing from SKILL.md, where indeed Small Language Models (SLMs) \cite{li-etal-2025-small} often exhibit suboptimal success rates and lower performance than larger counterparts in choosing the right skill \cite{belcak2025smalllanguagemodelsfuture, LI2025242}.

\section{Problem and Experiments}

\subsection{Research Focus}

When applying agent skills to complex tasks, LLMs read the SKILL.md from a ``skill hub'', which describes the next steps, how to use specific tools, and where to find additional guidance. This setup is well suited to ReAct-style agent systems \cite{yao2023reactsynergizingreasoningacting} that solve problems by thinking, taking an action, and adjusting based on the results, as shown in Figure~\ref{fig:skill-agent-framework}. First, the agent searches its skill library and selects the most appropriate skill for the task. It then plans, executes tools step by step, and adapts based on tool outputs or new data, and can even leverage an MCP server. After each step, the agent decides whether to continue, gather more information via MCP connections or other references, switch to a different skill, or terminate. This process helps prevent context explosion and enables more efficient use of \textbf{context windows}, which are among the most precious resources in agentic frameworks \cite{liu2026diveclaudecodedesign}. As illustrated, the \textbf{core capability} lies in selecting appropriate skills from a skill hub and subsequently performing planning, decision-making, and execution based on the selected skill content. Accordingly, this paper focuses on whether smaller-scale models can benefit from skills, how large the resulting performance gains are, and what the associated GPU costs are.

\subsection{Methods}

To evaluate skill-enabled agents, a temporary skill repository is constructed for each task by sampling 5–8 distractor skill entries from a publicly collected skill hub and combining them with the ground-truth skill (later extended to up to 100 skills). Then, we assess how three CE strategies affect agent performance and efficiency in complex decision-making. (1) \textbf{Direct Instruction (DI)} uses a minimal prompt to mimic raw user input. (2) \textbf{Full-Skill Instruction (FSI)} provides a fixed context containing the entire temporary Skill repository, requiring the model to identify the correct skill among multiple specifications. (3) \textbf{Agent Skill Instruction (ASI)} loads skill information on demand: the model determines whether additional skill details are needed, retrieves the relevant skill, and generates an answer conditioned on that information. If additional skill information is required, the relevant skill is loaded into the context, and the model makes a decision to use external tools to generate an answer conditioned on that information.

\subsection{Datasets}

\begin{table}[htbp]
\centering
\caption{Dataset overview for the evaluations: average input length (words/items), number of classes where applicable, domain/topic, and evaluation set size.}
\label{tab:dataset_stats}
\begin{adjustbox}{max width=1.0\linewidth}
\begin{tabular}{l c c l c c}
\toprule
\textbf{Dataset} & \textbf{Avg.\ Len} & \textbf{\# Cls} & \textbf{Topic / Domain} & \textbf{\# Eval} & \textbf{Ref.} \\ 
\midrule
IMDB       & 74.05  & 2   & Film reviews     & 300 & \citeauthor{maas-EtAl:2011:ACL-HLT2011} \\
FiNER      & 50.43  & 139 & Financial tags   & 403 & \citeauthor{loukas-etal-2022-finer} \\
InsurBench & 710.52 & 2   & Insurance claims & 300 & -- \\
MBPP       & --     & --  & Python Coding & 300 & \citeauthor{austin2021programsynthesislargelanguage} \\
\bottomrule
\end{tabular}
\end{adjustbox}
\end{table}

As shown in Table \ref{tab:dataset_stats}, to evaluate model-selection strategies in this setting, we design a benchmark suite with four datasets that span increasing task complexity and tool dependence: \textbf{IMDB} serves as a simple sanity check to ensure skill augmentation does not degrade basic classification capabilities~\cite{maas-EtAl:2011:ACL-HLT2011}; \textbf{FiNER} tests domain-specific instructions in a financial tagging task with 139 tag types~\cite{loukas-etal-2022-finer}; \textbf{InsurBench}, built from authentic insurance-claim email histories, is the primary industrial case study for long-context, noisy decision-making under proprietary constraints and multilingual conversations, where with SLMs achieve accuracy less than 40\% without fine-tuning; and \textbf{MBPP} provides exploratory validation of tool calling and ReAct-style behavior in coding agent settings such as Claude Code or Codex \cite{austin2021programsynthesislargelanguage} under agent skill settings. Our experiments are motivated by an industrial deployment scenario: an in-house insurance-claim assistant running on a private GPU cluster without external API access, where business processes, compliance rules, and tool instructions are encoded in a skill repository. Under tight GPU budgets, deployment teams must trade off accuracy, latency, and VRAM usage.

\begin{table}[htbp]
\centering
\caption{Model inventory. VRAM values correspond to approximate inference memory requirements under our deployment configuration.}
\label{tab:model_overview}
\begin{adjustbox}{max width=1.0\linewidth}
\begin{tabular}{l c c c}
\toprule
\textbf{Model} & \textbf{Size} & \textbf{VRAM (GB)} & \textbf{Release} \\
\midrule
\texttt{gpt-4o-mini}                      & --     & --   & 07/2024 \cite{openai_gpt4omini_docs}\\
\texttt{gemma-3-270m-it}                  & 0.27B  & 1    & 07/2025 \cite{huggingface_gemma3_270m_it} \\
\texttt{gemma-3-4b-it}                    & 4B     & 10   & 03/2025 \cite{huggingface_gemma3_4b_it} \\
\texttt{gemma-3-12b-it}                   & 12B    & 29   & 03/2025 \cite{huggingface_gemma3_12b_it}\\
\texttt{Qwen3-30B-A3B-Instruct}           & 30B    & 72   & 07/2025 \cite{huggingface_qwen3_30b_a3b_instruct_2507}\\
\texttt{Qwen3-Next-80B-A3B-Instruct}      & 80B    & 192  & 09/2025 \cite{huggingface_qwen3_next_80b_a3b_instruct}\\
\texttt{Qwen3-Next-80B-A3B-Thinking}      & 80B    & 192  & 09/2025 \cite{huggingface_qwen3_next_80b_a3b_thinking}\\
\texttt{Qwen3-Coder-Next}                 & 80B    & 192  & 01/2026 \cite{qwen_qwen3_coder_next_tech_report}\\
\bottomrule
\end{tabular}
\end{adjustbox}
\end{table}

\subsection{Small And Medium Language Models}

What qualifies as a ``small'' model remains debatable. We adopt the definition that the upper bound for large- and medium-sized models should be determined by the target task~\cite{wang2024comprehensivesurveysmalllanguage} --- in agentic contexts, models at the scale of Claude Sonnet and Fable are considered large, potentially exceeding 1T parameters. Following this convention, we treat models with fewer than 30B parameters as SLMs throughout this paper, and also include 80B models as a medium-sized comparison point. We also include different variants at the same scale. In this work, \textit{code} variants of LLMs are specialized for programming-centric tasks such as code generation, completion, and repair, whereas ``thinking'' models are optimized for multi-step deliberation, planning, and verification, and tend to excel at tasks requiring decomposition and derivation. Accordingly, our evaluations are restricted to open-source models spanning a wide range of scales, from 270M to 80B parameters, including one closed-source model, GPT-4o-mini, as shown in Table~\ref{tab:model_overview}. Each experiment is repeated five times with different random seeds to ensure statistically reproducible and significant results.

\subsection{Experimental Settings}

The experiments are structured to emphasize two core aspects: 
(i) \emph{skill selection}: selecting (routing to) the appropriate skill for a given task, and 
(ii) the agent's execution correctness after the selected skill is selected, thereby assessing the validity of the skill itself. For evaluation, \textbf{Cls ACC (classification accuracy)} and \textbf{Cls F1 (F1 score)} quantify classification performance, together with \textbf{Skill ACC (skill-selection accuracy)} for routing quality.  To reflect practical efficiency in industrial settings, we report \textbf{Avg GT (min)}, the average GPU processing time per task in minutes, capturing wall-clock latency. We further introduce \textbf{Avg VRAM Time (GB$\cdot$min)}, defined as the product of average reserved GPU memory and processing time per task, which quantifies the memory--time cost. This design approximates common GPU-cluster cost models, where operational cost scales with both wall-clock latency and sustained memory residency, as longer jobs and higher VRAM occupancy reduce cluster throughput. Moreover, under fixed VRAM budgets, memory occupancy can become a primary throughput bottleneck: once GPU memory is saturated, concurrent jobs are blocked, an effect not captured by compute-centric metrics such as FLOPS.


\section{Results}

\begin{table*}[htbp]
\centering
\caption{Main performance on IMDB, FiNER, InsurBench, and MBPP (robust parser accuracy). Skill ACC applies only to ASI; Avg GT (min) and Avg VRAM Time (GB$\cdot$min) are per-task GPU time and GPU-memory--time. Superscript/subscript values denote 95\% Wald confidence intervals for accuracy and skill-selection accuracy, truncated to [0,1] when necessary. gpt-4o-mini is a single-run baseline (blank on these datasets). For MBPP, \textbf{no Cls F1 is reported} because it is a coding task evaluated solely by test-case pass rate, for which F1 is not applicable.}
\label{tab:bench_full}
\begin{adjustbox}{max width=1.00\textwidth,center}
\small
\setlength{\tabcolsep}{3.2pt}
\renewcommand{\arraystretch}{1.40}

\begin{tabular}{ll *{5}{c} *{5}{c} *{5}{c} *{5}{c}}
\toprule
\multirow{3}{*}{Model Name} & \multirow{3}{*}{Method}
& \multicolumn{5}{c}{IMDB}
& \multicolumn{5}{c}{FiNER}
& \multicolumn{5}{c}{InsurBench}
& \multicolumn{5}{c}{MBPP} \\
\cmidrule(lr){3-7}\cmidrule(lr){8-12}\cmidrule(lr){13-17}\cmidrule(lr){18-22}
& 
& Cls & Cls & Skill & AVG & VRAM $\times$
& Cls & Cls & Skill & AVG & VRAM $\times$
& Cls & Cls & Skill & AVG & VRAM $\times$
& Pass & Cls & Skill & AVG & VRAM $\times$ \\
&
& ACC.(\(\uparrow\)) & F1(\(\uparrow\)) & ACC.(\(\uparrow\)) & GT(\(\downarrow\)) & Time (\(\downarrow\))
& ACC.(\(\uparrow\)) & F1(\(\uparrow\)) & ACC.(\(\uparrow\)) & GT(\(\downarrow\)) & Time (\(\downarrow\))
& ACC.(\(\uparrow\)) & F1(\(\uparrow\)) & ACC.(\(\uparrow\)) & GT(\(\downarrow\)) & Time (\(\downarrow\)) & ACC.(\(\uparrow\)) & F1(\(\uparrow\)) & ACC.(\(\uparrow\)) & GT(\(\downarrow\)) & Time(\(\downarrow\))\\

\midrule

\multirow{1}{*}{gpt-4o-mini (baseline)} & DI
& 0.85 & 0.60 & -- & -- & --
& 0.48 & 0.30 & -- & -- & --
& -- & -- & -- & -- & --
& 0.84 & -- & -- & -- & -- \\
\midrule
\midrule

\multirow{3}{*}{Qwen3-Next-80B-A3B-Thinking} & DI
& 0.69$^{\scriptscriptstyle 0.67}_{\scriptscriptstyle 0.71}$ & 0.53$^{\scriptscriptstyle 0.53}_{\scriptscriptstyle 0.54}$ & -- & \textbf{0.15}$^{\scriptscriptstyle 0.15}_{\scriptscriptstyle 0.15}$ & \textbf{24.93}$^{\scriptscriptstyle 24.51}_{\scriptscriptstyle 25.36}$
& 0.62$^{\scriptscriptstyle 0.61}_{\scriptscriptstyle 0.63}$ & 0.48$^{\scriptscriptstyle 0.46}_{\scriptscriptstyle 0.51}$ & -- & 0.30$^{\scriptscriptstyle 0.25}_{\scriptscriptstyle 0.36}$ & 50.31$^{\scriptscriptstyle 41.73}_{\scriptscriptstyle 58.89}$
& 0.54$^{\scriptscriptstyle 0.47}_{\scriptscriptstyle 0.61}$ & \textbf{0.33} & -- & 0.35$^{\scriptscriptstyle 0.34}_{\scriptscriptstyle 0.37}$ & 58.61$^{\scriptscriptstyle 55.44}_{\scriptscriptstyle 61.77}$
& 0.26$^{\scriptscriptstyle 0.24}_{\scriptscriptstyle 0.28}$ & -- & -- & \textbf{0.18}$^{\scriptscriptstyle 0.17}_{\scriptscriptstyle 0.19}$ & \textbf{30.13}$^{\scriptscriptstyle 28.90}_{\scriptscriptstyle 31.37}$ \\
& FSI
& 0.78$^{\scriptscriptstyle 0.76}_{\scriptscriptstyle 0.80}$ & 0.57$^{\scriptscriptstyle 0.56}_{\scriptscriptstyle 0.59}$ & -- & 0.19$^{\scriptscriptstyle 0.18}_{\scriptscriptstyle 0.19}$ & 30.84$^{\scriptscriptstyle 30.48}_{\scriptscriptstyle 31.21}$
& 0.55$^{\scriptscriptstyle 0.53}_{\scriptscriptstyle 0.57}$ & 0.44$^{\scriptscriptstyle 0.42}_{\scriptscriptstyle 0.47}$ & -- & \textbf{0.23}$^{\scriptscriptstyle 0.19}_{\scriptscriptstyle 0.27}$ & \textbf{37.64}$^{\scriptscriptstyle 31.49}_{\scriptscriptstyle 43.78}$
& 0.49$^{\scriptscriptstyle 0.42}_{\scriptscriptstyle 0.56}$ & 0.31 & -- & \textbf{0.28}$^{\scriptscriptstyle 0.26}_{\scriptscriptstyle 0.31}$ & \textbf{46.81}$^{\scriptscriptstyle 43.12}_{\scriptscriptstyle 50.50}$
& 0.22$^{\scriptscriptstyle 0.21}_{\scriptscriptstyle 0.23}$ & -- & -- & 0.19$^{\scriptscriptstyle 0.18}_{\scriptscriptstyle 0.20}$ & 31.12$^{\scriptscriptstyle 29.73}_{\scriptscriptstyle 32.51}$ \\
\rowcolor{asblue}
& ASI
& \textbf{0.97}$^{\scriptscriptstyle 0.96}_{\scriptscriptstyle 0.97}$ & \textbf{0.97}$^{\scriptscriptstyle 0.96}_{\scriptscriptstyle 0.97}$ & \textbf{1.00}$^{\scriptscriptstyle 1.00}_{\scriptscriptstyle 1.00}$ & 0.30$^{\scriptscriptstyle 0.30}_{\scriptscriptstyle 0.31}$ & 50.13$^{\scriptscriptstyle 49.62}_{\scriptscriptstyle 50.63}$
& \textbf{0.68}$^{\scriptscriptstyle 0.66}_{\scriptscriptstyle 0.69}$ & \textbf{0.53}$^{\scriptscriptstyle 0.52}_{\scriptscriptstyle 0.54}$ & \textbf{0.91}$^{\scriptscriptstyle 0.90}_{\scriptscriptstyle 0.92}$ & 0.56$^{\scriptscriptstyle 0.47}_{\scriptscriptstyle 0.65}$ & 92.40$^{\scriptscriptstyle 77.67}_{\scriptscriptstyle 107.12}$
& \textbf{0.55}$^{\scriptscriptstyle 0.48}_{\scriptscriptstyle 0.61}$ & \textbf{0.33} & \textbf{0.92}$^{\scriptscriptstyle 0.88}_{\scriptscriptstyle 0.96}$ & 0.55$^{\scriptscriptstyle 0.52}_{\scriptscriptstyle 0.58}$ & 90.70$^{\scriptscriptstyle 85.31}_{\scriptscriptstyle 96.09}$
& \textbf{0.71}$^{\scriptscriptstyle 0.71}_{\scriptscriptstyle 0.72}$ & -- & \textbf{1.00}$^{\scriptscriptstyle 1.00}_{\scriptscriptstyle 1.00}$ & 1.52$^{\scriptscriptstyle 1.45}_{\scriptscriptstyle 1.59}$ & 251.36$^{\scriptscriptstyle 239.99}_{\scriptscriptstyle 262.72}$ \\
\midrule

\multirow{3}{*}{Qwen3-Next-80B-A3B-Instruct} & DI
& 0.77$^{\scriptscriptstyle 0.76}_{\scriptscriptstyle 0.79}$ & 0.57$^{\scriptscriptstyle 0.56}_{\scriptscriptstyle 0.58}$ & -- & 0.05$^{\scriptscriptstyle 0.05}_{\scriptscriptstyle 0.06}$ & 8.47$^{\scriptscriptstyle 7.91}_{\scriptscriptstyle 9.02}$
& 0.61$^{\scriptscriptstyle 0.59}_{\scriptscriptstyle 0.62}$ & \textbf{0.48}$^{\scriptscriptstyle 0.46}_{\scriptscriptstyle 0.49}$ & -- & 0.09$^{\scriptscriptstyle 0.07}_{\scriptscriptstyle 0.10}$ & 14.52$^{\scriptscriptstyle 11.93}_{\scriptscriptstyle 17.11}$
& 0.59$^{\scriptscriptstyle 0.57}_{\scriptscriptstyle 0.61}$ & 0.40$^{\scriptscriptstyle 0.38}_{\scriptscriptstyle 0.41}$ & -- & 0.13$^{\scriptscriptstyle 0.11}_{\scriptscriptstyle 0.14}$ & 20.69$^{\scriptscriptstyle 17.76}_{\scriptscriptstyle 23.62}$
& 0.88$^{\scriptscriptstyle 0.87}_{\scriptscriptstyle 0.89}$ & -- & -- & \textbf{0.02}$^{\scriptscriptstyle 0.02}_{\scriptscriptstyle 0.02}$ & \textbf{3.61}$^{\scriptscriptstyle 3.48}_{\scriptscriptstyle 3.74}$ \\
& FSI
& 0.72$^{\scriptscriptstyle 0.70}_{\scriptscriptstyle 0.74}$ & 0.55$^{\scriptscriptstyle 0.54}_{\scriptscriptstyle 0.56}$ & -- & \textbf{0.04}$^{\scriptscriptstyle 0.04}_{\scriptscriptstyle 0.05}$ & \textbf{7.07}$^{\scriptscriptstyle 6.65}_{\scriptscriptstyle 7.49}$
& 0.61$^{\scriptscriptstyle 0.60}_{\scriptscriptstyle 0.62}$ & 0.45$^{\scriptscriptstyle 0.43}_{\scriptscriptstyle 0.46}$ & -- & 0.08$^{\scriptscriptstyle 0.06}_{\scriptscriptstyle 0.09}$ & 12.68$^{\scriptscriptstyle 10.36}_{\scriptscriptstyle 15.01}$
& 0.34$^{\scriptscriptstyle 0.33}_{\scriptscriptstyle 0.36}$ & 0.21$^{\scriptscriptstyle 0.21}_{\scriptscriptstyle 0.22}$ & -- & \textbf{0.05}$^{\scriptscriptstyle 0.05}_{\scriptscriptstyle 0.06}$ & \textbf{8.62}$^{\scriptscriptstyle 7.44}_{\scriptscriptstyle 9.80}$
& 0.88$^{\scriptscriptstyle 0.88}_{\scriptscriptstyle 0.89}$ & -- & -- & \textbf{0.02}$^{\scriptscriptstyle 0.02}_{\scriptscriptstyle 0.02}$ & 3.68$^{\scriptscriptstyle 3.58}_{\scriptscriptstyle 3.78}$ \\
\rowcolor{asblue}
& ASI
& \textbf{0.93}$^{\scriptscriptstyle 0.92}_{\scriptscriptstyle 0.93}$ & \textbf{0.62}$^{\scriptscriptstyle 0.62}_{\scriptscriptstyle 0.62}$ & \textbf{0.98}$^{\scriptscriptstyle 0.98}_{\scriptscriptstyle 0.99}$ & \textbf{0.04}$^{\scriptscriptstyle 0.04}_{\scriptscriptstyle 0.05}$ & 7.31$^{\scriptscriptstyle 6.90}_{\scriptscriptstyle 7.73}$
& \textbf{0.63}$^{\scriptscriptstyle 0.62}_{\scriptscriptstyle 0.64}$ & \textbf{0.48}$^{\scriptscriptstyle 0.46}_{\scriptscriptstyle 0.50}$ & \textbf{0.98}$^{\scriptscriptstyle 0.98}_{\scriptscriptstyle 0.98}$ & \textbf{0.06}$^{\scriptscriptstyle 0.05}_{\scriptscriptstyle 0.07}$ & \textbf{10.05}$^{\scriptscriptstyle 8.33}_{\scriptscriptstyle 11.76}$
& \textbf{0.63}$^{\scriptscriptstyle 0.62}_{\scriptscriptstyle 0.64}$ & \textbf{0.47}$^{\scriptscriptstyle 0.46}_{\scriptscriptstyle 0.48}$ & \textbf{0.93}$^{\scriptscriptstyle 0.93}_{\scriptscriptstyle 0.94}$ & 0.06$^{\scriptscriptstyle 0.06}_{\scriptscriptstyle 0.07}$ & 10.58$^{\scriptscriptstyle 9.18}_{\scriptscriptstyle 11.97}$
& \textbf{0.92}$^{\scriptscriptstyle 0.92}_{\scriptscriptstyle 0.93}$ & -- & \textbf{1.00}$^{\scriptscriptstyle 1.00}_{\scriptscriptstyle 1.00}$ & 0.03$^{\scriptscriptstyle 0.03}_{\scriptscriptstyle 0.03}$ & 5.34$^{\scriptscriptstyle 5.05}_{\scriptscriptstyle 5.64}$ \\
\midrule

\multirow{3}{*}{Qwen3-Coder-Next} & DI
& 0.83$^{\scriptscriptstyle 0.82}_{\scriptscriptstyle 0.84}$ & 0.59$^{\scriptscriptstyle 0.59}_{\scriptscriptstyle 0.60}$ & -- & \textbf{0.05}$^{\scriptscriptstyle 0.05}_{\scriptscriptstyle 0.06}$ & 8.50$^{\scriptscriptstyle 7.83}_{\scriptscriptstyle 9.17}$
& 0.46$^{\scriptscriptstyle 0.45}_{\scriptscriptstyle 0.47}$ & 0.37$^{\scriptscriptstyle 0.35}_{\scriptscriptstyle 0.38}$ & -- & 0.08$^{\scriptscriptstyle 0.06}_{\scriptscriptstyle 0.10}$ & 12.90$^{\scriptscriptstyle 10.12}_{\scriptscriptstyle 15.67}$
& \textbf{0.63}$^{\scriptscriptstyle 0.60}_{\scriptscriptstyle 0.65}$ & \textbf{0.55}$^{\scriptscriptstyle 0.43}_{\scriptscriptstyle 0.66}$ & -- & 0.13$^{\scriptscriptstyle 0.11}_{\scriptscriptstyle 0.15}$ & 21.42$^{\scriptscriptstyle 17.64}_{\scriptscriptstyle 25.20}$
& 0.47$^{\scriptscriptstyle 0.46}_{\scriptscriptstyle 0.49}$ & -- & -- & \textbf{0.03}$^{\scriptscriptstyle 0.03}_{\scriptscriptstyle 0.04}$ & \textbf{5.14}$^{\scriptscriptstyle 4.25}_{\scriptscriptstyle 6.04}$ \\
& FSI
& 0.62$^{\scriptscriptstyle 0.61}_{\scriptscriptstyle 0.64}$ & 0.50$^{\scriptscriptstyle 0.50}_{\scriptscriptstyle 0.51}$ & -- & \textbf{0.05}$^{\scriptscriptstyle 0.04}_{\scriptscriptstyle 0.05}$ & \textbf{7.74}$^{\scriptscriptstyle 7.24}_{\scriptscriptstyle 8.24}$
& 0.47$^{\scriptscriptstyle 0.45}_{\scriptscriptstyle 0.49}$ & 0.36$^{\scriptscriptstyle 0.35}_{\scriptscriptstyle 0.37}$ & -- & 0.07$^{\scriptscriptstyle 0.06}_{\scriptscriptstyle 0.09}$ & 11.97$^{\scriptscriptstyle 9.58}_{\scriptscriptstyle 14.36}$
& 0.47$^{\scriptscriptstyle 0.43}_{\scriptscriptstyle 0.52}$ & 0.36$^{\scriptscriptstyle 0.33}_{\scriptscriptstyle 0.39}$ & -- & \textbf{0.06}$^{\scriptscriptstyle 0.05}_{\scriptscriptstyle 0.07}$ & \textbf{9.69}$^{\scriptscriptstyle 8.08}_{\scriptscriptstyle 11.31}$
& 0.86$^{\scriptscriptstyle 0.85}_{\scriptscriptstyle 0.87}$ & -- & -- & 0.05$^{\scriptscriptstyle 0.04}_{\scriptscriptstyle 0.06}$ & 8.71$^{\scriptscriptstyle 7.05}_{\scriptscriptstyle 10.37}$ \\
\rowcolor{asblue}
& ASI
& \textbf{0.89}$^{\scriptscriptstyle 0.87}_{\scriptscriptstyle 0.91}$ & \textbf{0.62}$^{\scriptscriptstyle 0.61}_{\scriptscriptstyle 0.62}$ & \textbf{1.00}$^{\scriptscriptstyle 1.00}_{\scriptscriptstyle 1.00}$ & \textbf{0.05}$^{\scriptscriptstyle 0.05}_{\scriptscriptstyle 0.05}$ & 8.15$^{\scriptscriptstyle 7.73}_{\scriptscriptstyle 8.56}$
& \textbf{0.56}$^{\scriptscriptstyle 0.54}_{\scriptscriptstyle 0.58}$ & \textbf{0.41}$^{\scriptscriptstyle 0.39}_{\scriptscriptstyle 0.42}$ & \textbf{0.99}$^{\scriptscriptstyle 0.99}_{\scriptscriptstyle 0.99}$ & \textbf{0.05}$^{\scriptscriptstyle 0.04}_{\scriptscriptstyle 0.06}$ & \textbf{8.12}$^{\scriptscriptstyle 6.70}_{\scriptscriptstyle 9.54}$
& 0.49$^{\scriptscriptstyle 0.46}_{\scriptscriptstyle 0.52}$ & 0.39$^{\scriptscriptstyle 0.36}_{\scriptscriptstyle 0.41}$ & \textbf{0.97}$^{\scriptscriptstyle 0.97}_{\scriptscriptstyle 0.98}$ & 0.10$^{\scriptscriptstyle 0.08}_{\scriptscriptstyle 0.12}$ & 16.36$^{\scriptscriptstyle 13.53}_{\scriptscriptstyle 19.20}$
& \textbf{0.94}$^{\scriptscriptstyle 0.94}_{\scriptscriptstyle 0.95}$ & -- & \textbf{1.00}$^{\scriptscriptstyle 1.00}_{\scriptscriptstyle 1.00}$ & 0.09$^{\scriptscriptstyle 0.08}_{\scriptscriptstyle 0.10}$ & 14.92$^{\scriptscriptstyle 12.75}_{\scriptscriptstyle 17.08}$ \\
\midrule

\multirow{3}{*}{Qwen3-30B-A3B-Instruct}
& DI
& 0.65$^{\scriptscriptstyle 0.60}_{\scriptscriptstyle 0.70}$ & 0.46$^{\scriptscriptstyle 0.35}_{\scriptscriptstyle 0.57}$ & -- & \textbf{0.19}$^{\scriptscriptstyle 0.11}_{\scriptscriptstyle 0.27}$ & 12.38$^{\scriptscriptstyle 7.20}_{\scriptscriptstyle 17.56}$
& 0.28$^{\scriptscriptstyle 0.25}_{\scriptscriptstyle 0.31}$ & 0.20$^{\scriptscriptstyle 0.18}_{\scriptscriptstyle 0.23}$ & -- & \textbf{0.15}$^{\scriptscriptstyle 0.05}_{\scriptscriptstyle 0.25}$ & \textbf{9.49}$^{\scriptscriptstyle 2.96}_{\scriptscriptstyle 16.02}$
& 0.23$^{\scriptscriptstyle 0.17}_{\scriptscriptstyle 0.29}$ & 0.19 & -- & 0.27$^{\scriptscriptstyle 0.22}_{\scriptscriptstyle 0.32}$ & 17.39$^{\scriptscriptstyle 14.33}_{\scriptscriptstyle 20.45}$
& 0.86$^{\scriptscriptstyle 0.86}_{\scriptscriptstyle 0.87}$ & -- & -- & \textbf{0.01}$^{\scriptscriptstyle 0.01}_{\scriptscriptstyle 0.01}$ & \textbf{0.69}$^{\scriptscriptstyle 0.55}_{\scriptscriptstyle 0.83}$ \\
& FSI
& 0.66$^{\scriptscriptstyle 0.61}_{\scriptscriptstyle 0.70}$ & 0.47$^{\scriptscriptstyle 0.36}_{\scriptscriptstyle 0.58}$ & -- & 0.20$^{\scriptscriptstyle 0.11}_{\scriptscriptstyle 0.28}$ & 12.73$^{\scriptscriptstyle 7.25}_{\scriptscriptstyle 18.22}$
& 0.27$^{\scriptscriptstyle 0.24}_{\scriptscriptstyle 0.30}$ & 0.19$^{\scriptscriptstyle 0.16}_{\scriptscriptstyle 0.22}$ & -- & 0.16$^{\scriptscriptstyle 0.06}_{\scriptscriptstyle 0.25}$ & 10.05$^{\scriptscriptstyle 3.61}_{\scriptscriptstyle 16.48}$
& 0.21$^{\scriptscriptstyle 0.15}_{\scriptscriptstyle 0.27}$ & 0.17 & -- & \textbf{0.22}$^{\scriptscriptstyle 0.19}_{\scriptscriptstyle 0.25}$ & \textbf{14.28}$^{\scriptscriptstyle 12.16}_{\scriptscriptstyle 16.40}$
& 0.87$^{\scriptscriptstyle 0.86}_{\scriptscriptstyle 0.87}$ & -- & -- & \textbf{0.01}$^{\scriptscriptstyle 0.01}_{\scriptscriptstyle 0.01}$ & 0.73$^{\scriptscriptstyle 0.65}_{\scriptscriptstyle 0.82}$ \\
\rowcolor{asblue}
& ASI
& \textbf{0.94}$^{\scriptscriptstyle 0.92}_{\scriptscriptstyle 0.95}$ & \textbf{0.85}$^{\scriptscriptstyle 0.67}_{\scriptscriptstyle 1.00}$ & \textbf{1.00}$^{\scriptscriptstyle 1.00}_{\scriptscriptstyle 1.00}$ & \textbf{0.19}$^{\scriptscriptstyle 0.11}_{\scriptscriptstyle 0.27}$ & \textbf{12.21}$^{\scriptscriptstyle 6.87}_{\scriptscriptstyle 17.55}$
& \textbf{0.36}$^{\scriptscriptstyle 0.30}_{\scriptscriptstyle 0.43}$ & \textbf{0.30}$^{\scriptscriptstyle 0.26}_{\scriptscriptstyle 0.35}$ & \textbf{1.00}$^{\scriptscriptstyle 0.99}_{\scriptscriptstyle 1.00}$ & 0.55$^{\scriptscriptstyle 0.33}_{\scriptscriptstyle 0.77}$ & 35.81$^{\scriptscriptstyle 21.28}_{\scriptscriptstyle 50.34}$
& \textbf{0.30}$^{\scriptscriptstyle 0.24}_{\scriptscriptstyle 0.36}$ & \textbf{0.28} & \textbf{0.99}$^{\scriptscriptstyle 0.98}_{\scriptscriptstyle 1.00}$ & 0.40$^{\scriptscriptstyle 0.34}_{\scriptscriptstyle 0.46}$ & 26.09$^{\scriptscriptstyle 22.15}_{\scriptscriptstyle 30.04}$
& \textbf{0.89}$^{\scriptscriptstyle 0.88}_{\scriptscriptstyle 0.89}$ & -- & \textbf{1.00}$^{\scriptscriptstyle 1.00}_{\scriptscriptstyle 1.00}$ & 0.03$^{\scriptscriptstyle 0.03}_{\scriptscriptstyle 0.03}$ & 2.02$^{\scriptscriptstyle 1.81}_{\scriptscriptstyle 2.22}$ \\
\midrule

\multirow{3}{*}{gemma-3-12b-it} & DI
& \textbf{0.71}$^{\scriptscriptstyle 0.51}_{\scriptscriptstyle 0.92}$ & 0.52$^{\scriptscriptstyle 0.38}_{\scriptscriptstyle 0.66}$ & -- & 0.08$^{\scriptscriptstyle 0.05}_{\scriptscriptstyle 0.12}$ & 2.35$^{\scriptscriptstyle 1.34}_{\scriptscriptstyle 3.37}$
& \textbf{0.47}$^{\scriptscriptstyle 0.44}_{\scriptscriptstyle 0.51}$ & 0.33$^{\scriptscriptstyle 0.30}_{\scriptscriptstyle 0.37}$ & -- & 0.07$^{\scriptscriptstyle 0.05}_{\scriptscriptstyle 0.08}$ & 1.82$^{\scriptscriptstyle 1.41}_{\scriptscriptstyle 2.23}$
& 0.55$^{\scriptscriptstyle 0.52}_{\scriptscriptstyle 0.59}$ & 0.35$^{\scriptscriptstyle 0.33}_{\scriptscriptstyle 0.37}$ & -- & 0.09$^{\scriptscriptstyle 0.07}_{\scriptscriptstyle 0.12}$ & 2.63$^{\scriptscriptstyle 1.98}_{\scriptscriptstyle 3.28}$
& \textbf{0.85}$^{\scriptscriptstyle 0.84}_{\scriptscriptstyle 0.85}$ & -- & -- & \textbf{0.02}$^{\scriptscriptstyle 0.02}_{\scriptscriptstyle 0.03}$ & 0.67$^{\scriptscriptstyle 0.63}_{\scriptscriptstyle 0.70}$ \\
& FSI
& 0.63$^{\scriptscriptstyle 0.44}_{\scriptscriptstyle 0.82}$ & \textbf{0.57}$^{\scriptscriptstyle 0.35}_{\scriptscriptstyle 0.78}$ & -- & 0.16$^{\scriptscriptstyle 0.09}_{\scriptscriptstyle 0.22}$ & 4.35$^{\scriptscriptstyle 2.49}_{\scriptscriptstyle 6.21}$
& 0.22$^{\scriptscriptstyle 0.18}_{\scriptscriptstyle 0.26}$ & 0.19$^{\scriptscriptstyle 0.15}_{\scriptscriptstyle 0.22}$ & -- & 0.12$^{\scriptscriptstyle 0.08}_{\scriptscriptstyle 0.15}$ & 3.21$^{\scriptscriptstyle 2.35}_{\scriptscriptstyle 4.08}$
& 0.50$^{\scriptscriptstyle 0.48}_{\scriptscriptstyle 0.52}$ & 0.33$^{\scriptscriptstyle 0.31}_{\scriptscriptstyle 0.34}$ & -- & 0.11$^{\scriptscriptstyle 0.08}_{\scriptscriptstyle 0.14}$ & 3.08$^{\scriptscriptstyle 2.24}_{\scriptscriptstyle 3.93}$
& 0.82$^{\scriptscriptstyle 0.82}_{\scriptscriptstyle 0.83}$ & -- & -- & \textbf{0.02}$^{\scriptscriptstyle 0.02}_{\scriptscriptstyle 0.02}$ & \textbf{0.64}$^{\scriptscriptstyle 0.62}_{\scriptscriptstyle 0.65}$ \\
\rowcolor{asblue}
& ASI
& 0.70$^{\scriptscriptstyle 0.55}_{\scriptscriptstyle 0.84}$ & 0.51$^{\scriptscriptstyle 0.41}_{\scriptscriptstyle 0.61}$ & \textbf{1.00}$^{\scriptscriptstyle 1.00}_{\scriptscriptstyle 1.00}$ & \textbf{0.07}$^{\scriptscriptstyle 0.05}_{\scriptscriptstyle 0.09}$ & \textbf{1.92}$^{\scriptscriptstyle 1.38}_{\scriptscriptstyle 2.46}$
& \textbf{0.47}$^{\scriptscriptstyle 0.45}_{\scriptscriptstyle 0.48}$ & \textbf{0.36}$^{\scriptscriptstyle 0.33}_{\scriptscriptstyle 0.39}$ & \textbf{1.00}$^{\scriptscriptstyle 1.00}_{\scriptscriptstyle 1.00}$ & \textbf{0.04}$^{\scriptscriptstyle 0.03}_{\scriptscriptstyle 0.05}$ & \textbf{1.01}$^{\scriptscriptstyle 0.75}_{\scriptscriptstyle 1.26}$
& \textbf{0.56}$^{\scriptscriptstyle 0.54}_{\scriptscriptstyle 0.57}$ & \textbf{0.43}$^{\scriptscriptstyle 0.34}_{\scriptscriptstyle 0.51}$ & \textbf{1.00}$^{\scriptscriptstyle 1.00}_{\scriptscriptstyle 1.00}$ & \textbf{0.08}$^{\scriptscriptstyle 0.06}_{\scriptscriptstyle 0.10}$ & \textbf{2.27}$^{\scriptscriptstyle 1.77}_{\scriptscriptstyle 2.77}$
& 0.82$^{\scriptscriptstyle 0.81}_{\scriptscriptstyle 0.82}$ & -- & \textbf{1.00}$^{\scriptscriptstyle 1.00}_{\scriptscriptstyle 1.00}$ & 0.05$^{\scriptscriptstyle 0.05}_{\scriptscriptstyle 0.05}$ & 1.43$^{\scriptscriptstyle 1.39}_{\scriptscriptstyle 1.47}$ \\
\midrule

\multirow{3}{*}{gemma-3-4b-it} & DI
& \textbf{0.79}$^{\scriptscriptstyle 0.76}_{\scriptscriptstyle 0.82}$ & \textbf{0.58}$^{\scriptscriptstyle 0.57}_{\scriptscriptstyle 0.59}$ & -- & 0.06$^{\scriptscriptstyle 0.05}_{\scriptscriptstyle 0.07}$ & 0.63$^{\scriptscriptstyle 0.54}_{\scriptscriptstyle 0.72}$
& \textbf{0.25}$^{\scriptscriptstyle 0.24}_{\scriptscriptstyle 0.26}$ & 0.14$^{\scriptscriptstyle 0.12}_{\scriptscriptstyle 0.16}$ & -- & \textbf{0.02}$^{\scriptscriptstyle 0.02}_{\scriptscriptstyle 0.03}$ & 0.24$^{\scriptscriptstyle 0.23}_{\scriptscriptstyle 0.26}$
& \textbf{0.47}$^{\scriptscriptstyle 0.40}_{\scriptscriptstyle 0.53}$ & \textbf{0.31}$^{\scriptscriptstyle 0.27}_{\scriptscriptstyle 0.36}$ & -- & 0.07$^{\scriptscriptstyle 0.07}_{\scriptscriptstyle 0.08}$ & 0.74$^{\scriptscriptstyle 0.73}_{\scriptscriptstyle 0.75}$
& \textbf{0.69}$^{\scriptscriptstyle 0.66}_{\scriptscriptstyle 0.71}$ & -- & -- & \textbf{0.02}$^{\scriptscriptstyle 0.01}_{\scriptscriptstyle 0.02}$ & \textbf{0.15}$^{\scriptscriptstyle 0.13}_{\scriptscriptstyle 0.16}$ \\
& FSI
& 0.58$^{\scriptscriptstyle 0.56}_{\scriptscriptstyle 0.60}$ & 0.48$^{\scriptscriptstyle 0.46}_{\scriptscriptstyle 0.49}$ & -- & 0.07$^{\scriptscriptstyle 0.06}_{\scriptscriptstyle 0.09}$ & 0.73$^{\scriptscriptstyle 0.62}_{\scriptscriptstyle 0.85}$
& 0.01$^{\scriptscriptstyle 0.00}_{\scriptscriptstyle 0.01}$ & 0.01$^{\scriptscriptstyle 0.00}_{\scriptscriptstyle 0.01}$ & -- & 0.08$^{\scriptscriptstyle 0.08}_{\scriptscriptstyle 0.08}$ & 0.80$^{\scriptscriptstyle 0.76}_{\scriptscriptstyle 0.84}$
& 0.34$^{\scriptscriptstyle 0.25}_{\scriptscriptstyle 0.42}$ & 0.26$^{\scriptscriptstyle 0.21}_{\scriptscriptstyle 0.31}$ & -- & 0.08$^{\scriptscriptstyle 0.08}_{\scriptscriptstyle 0.09}$ & 0.82$^{\scriptscriptstyle 0.77}_{\scriptscriptstyle 0.86}$
& 0.64$^{\scriptscriptstyle 0.62}_{\scriptscriptstyle 0.65}$ & -- & -- & \textbf{0.02}$^{\scriptscriptstyle 0.01}_{\scriptscriptstyle 0.02}$ & \textbf{0.15}$^{\scriptscriptstyle 0.14}_{\scriptscriptstyle 0.15}$ \\
\rowcolor{asblue}
& ASI
& 0.71$^{\scriptscriptstyle 0.36}_{\scriptscriptstyle 1.00}$ & 0.48$^{\scriptscriptstyle 0.25}_{\scriptscriptstyle 0.72}$ & \textbf{1.00}$^{\scriptscriptstyle 1.00}_{\scriptscriptstyle 1.00}$ & \textbf{0.03}$^{\scriptscriptstyle 0.02}_{\scriptscriptstyle 0.03}$ & \textbf{0.25}$^{\scriptscriptstyle 0.15}_{\scriptscriptstyle 0.34}$
& 0.21$^{\scriptscriptstyle 0.19}_{\scriptscriptstyle 0.22}$ & \textbf{0.15}$^{\scriptscriptstyle 0.13}_{\scriptscriptstyle 0.17}$ & \textbf{1.00}$^{\scriptscriptstyle 1.00}_{\scriptscriptstyle 1.00}$ & \textbf{0.02}$^{\scriptscriptstyle 0.02}_{\scriptscriptstyle 0.02}$ & \textbf{0.16}$^{\scriptscriptstyle 0.16}_{\scriptscriptstyle 0.17}$
& 0.08$^{\scriptscriptstyle 0.00}_{\scriptscriptstyle 0.19}$ & 0.07$^{\scriptscriptstyle 0.00}_{\scriptscriptstyle 0.17}$ & \textbf{0.98}$^{\scriptscriptstyle 0.96}_{\scriptscriptstyle 1.00}$ & \textbf{0.03}$^{\scriptscriptstyle 0.03}_{\scriptscriptstyle 0.03}$ & \textbf{0.33}$^{\scriptscriptstyle 0.31}_{\scriptscriptstyle 0.34}$
& 0.65$^{\scriptscriptstyle 0.64}_{\scriptscriptstyle 0.65}$ & -- & \textbf{1.00}$^{\scriptscriptstyle 1.00}_{\scriptscriptstyle 1.00}$ & 0.04$^{\scriptscriptstyle 0.04}_{\scriptscriptstyle 0.04}$ & 0.39$^{\scriptscriptstyle 0.38}_{\scriptscriptstyle 0.40}$ \\
\midrule

\multirow{3}{*}{gemma-3-270m-it} & DI
& \textbf{0.59}$^{\scriptscriptstyle 0.55}_{\scriptscriptstyle 0.62}$ & 0.38$^{\scriptscriptstyle 0.35}_{\scriptscriptstyle 0.41}$ & -- & \textbf{0.00}$^{\scriptscriptstyle 0.00}_{\scriptscriptstyle 0.00}$ & \textbf{0.00}$^{\scriptscriptstyle 0.00}_{\scriptscriptstyle 0.00}$
& 0.00$^{\scriptscriptstyle 0.00}_{\scriptscriptstyle 0.01}$ & 0.00$^{\scriptscriptstyle 0.00}_{\scriptscriptstyle 0.01}$ & -- & \textbf{0.01}$^{\scriptscriptstyle 0.01}_{\scriptscriptstyle 0.01}$ & \textbf{0.01}$^{\scriptscriptstyle 0.01}_{\scriptscriptstyle 0.01}$
& 0.06$^{\scriptscriptstyle 0.04}_{\scriptscriptstyle 0.08}$ & 0.06$^{\scriptscriptstyle 0.04}_{\scriptscriptstyle 0.09}$ & -- & \textbf{0.01}$^{\scriptscriptstyle 0.00}_{\scriptscriptstyle 0.01}$ & \textbf{0.01}$^{\scriptscriptstyle 0.01}_{\scriptscriptstyle 0.01}$
& \textbf{0.15}$^{\scriptscriptstyle 0.14}_{\scriptscriptstyle 0.16}$ & -- & -- & \textbf{0.01}$^{\scriptscriptstyle 0.00}_{\scriptscriptstyle 0.01}$ & \textbf{0.01}$^{\scriptscriptstyle 0.01}_{\scriptscriptstyle 0.01}$ \\
& FSI
& 0.02$^{\scriptscriptstyle 0.02}_{\scriptscriptstyle 0.03}$ & 0.03$^{\scriptscriptstyle 0.02}_{\scriptscriptstyle 0.04}$ & -- & 0.01$^{\scriptscriptstyle 0.01}_{\scriptscriptstyle 0.01}$ & 0.01$^{\scriptscriptstyle 0.01}_{\scriptscriptstyle 0.01}$
& 0.00$^{\scriptscriptstyle 0.00}_{\scriptscriptstyle 0.00}$ & 0.00$^{\scriptscriptstyle 0.00}_{\scriptscriptstyle 0.00}$ & -- & \textbf{0.01}$^{\scriptscriptstyle 0.01}_{\scriptscriptstyle 0.01}$ & \textbf{0.01}$^{\scriptscriptstyle 0.01}_{\scriptscriptstyle 0.01}$
& \textbf{0.10}$^{\scriptscriptstyle 0.09}_{\scriptscriptstyle 0.11}$ & \textbf{0.10}$^{\scriptscriptstyle 0.09}_{\scriptscriptstyle 0.11}$ & -- & \textbf{0.01}$^{\scriptscriptstyle 0.01}_{\scriptscriptstyle 0.01}$ & \textbf{0.01}$^{\scriptscriptstyle 0.01}_{\scriptscriptstyle 0.01}$
& 0.03$^{\scriptscriptstyle 0.02}_{\scriptscriptstyle 0.04}$ & -- & -- & 0.03$^{\scriptscriptstyle 0.02}_{\scriptscriptstyle 0.03}$ & 0.05$^{\scriptscriptstyle 0.05}_{\scriptscriptstyle 0.05}$ \\
\rowcolor{asblue}
& ASI
& 0.48$^{\scriptscriptstyle 0.24}_{\scriptscriptstyle 0.71}$ & \textbf{0.41}$^{\scriptscriptstyle 0.20}_{\scriptscriptstyle 0.62}$ & \textbf{0.04}$^{\scriptscriptstyle 0.00}_{\scriptscriptstyle 0.12}$ & \textbf{0.00}$^{\scriptscriptstyle 0.00}_{\scriptscriptstyle 0.00}$ & 0.01$^{\scriptscriptstyle 0.01}_{\scriptscriptstyle 0.01}$
& \textbf{0.02}$^{\scriptscriptstyle 0.01}_{\scriptscriptstyle 0.03}$ & \textbf{0.01}$^{\scriptscriptstyle 0.01}_{\scriptscriptstyle 0.01}$ & \textbf{0.02}$^{\scriptscriptstyle 0.00}_{\scriptscriptstyle 0.04}$ & \textbf{0.01}$^{\scriptscriptstyle 0.01}_{\scriptscriptstyle 0.01}$ & 0.02$^{\scriptscriptstyle 0.02}_{\scriptscriptstyle 0.02}$
& 0.02$^{\scriptscriptstyle 0.00}_{\scriptscriptstyle 0.03}$ & 0.02$^{\scriptscriptstyle 0.00}_{\scriptscriptstyle 0.04}$ & \textbf{0.10}$^{\scriptscriptstyle 0.04}_{\scriptscriptstyle 0.17}$ & 0.02$^{\scriptscriptstyle 0.01}_{\scriptscriptstyle 0.02}$ & 0.03$^{\scriptscriptstyle 0.02}_{\scriptscriptstyle 0.04}$
& 0.00$^{\scriptscriptstyle 0.00}_{\scriptscriptstyle 0.00}$ & -- & \textbf{0.12}$^{\scriptscriptstyle 0.18}_{\scriptscriptstyle 0.10}$ & 0.02$^{\scriptscriptstyle 0.02}_{\scriptscriptstyle 0.02}$ & 0.05$^{\scriptscriptstyle 0.05}_{\scriptscriptstyle 0.05}$ \\
\bottomrule

\end{tabular}
\end{adjustbox}
\end{table*}

\subsection{Main Performance}

\noindent\textbf{Skill Returns in SLMs} 
As shown in Table~\ref{tab:bench_full}, most SLMs exhibit clear performance improvements while maintaining high skill-selection accuracy. 
The gains are particularly pronounced for mid-sized and larger models; for example, on IMDB, Qwen3-Next-80B-A3B-Instruct improves Cls ACC from 0.77 to 0.93 and Qwen3-Next-80B-A3B-Thinking from 0.69 to 0.97 compared with Direct Instruction. 
In contrast, smaller models such as gemma-3-4b-it and gemma-3-270m-it show more limited or unstable improvements using agent skills. 
It is also observed that for simpler tasks (e.g., IMDB), the benefits of agent skills are modest for several models. 
For more challenging benchmarks such as FiNER and InsurBench, however, the results highlight the benefits of using skills in our evaluated settings. InsurBench further strengthens this conclusion, as its closed-source nature reduces the likelihood of dataset contamination.

\paragraph{Tiny Models Fail at Skill Routing} 
Across these datasets, each assessment includes distracting skills, whose presence is considered to further challenge the models' skill routing capabilities. Indeed, we find that extremely small models such as gemma-3-4b-it and gemma-3-270m-it still struggle to retrieve the appropriate skill. 
In particular, gemma-3-270m-it achieves near-zero Skill ACC on FiNER and InsurBench and only 0.04 on IMDB. Within an agent skill framework, models below 4B parameters often lack even the basic capability to reliably identify the correct skill, further compromising their capacity to carry out the subsequent, more complex execution steps.

\begin{figure*}[htbp]
    \centering
    \includegraphics[width=0.90\textwidth]{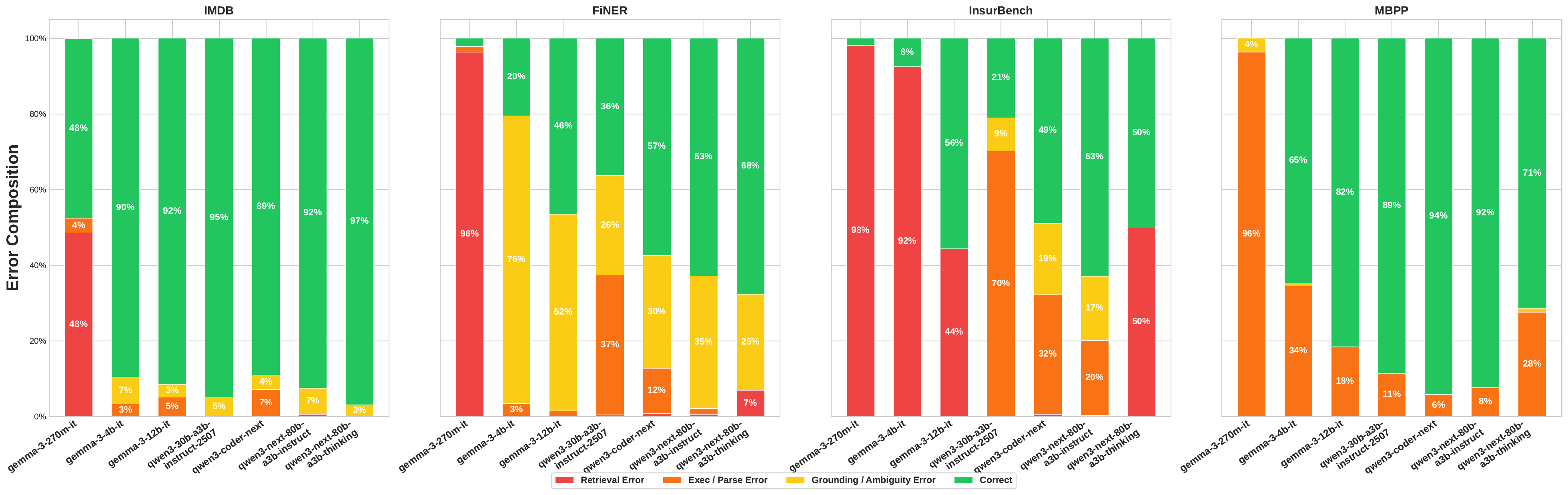}
    \caption{Error composition by model and benchmark. Each bar is normalized over all cases and decomposed into three failure stages: Retrieval Error (Wrong skill selected by agents), Exec / Parse Error (correct skill selected but output malformed), and Grounding / Ambiguity Error (correct skill and parsing, but incorrect final answer). Models are ordered by parameter size, with the Instruct, Coder, and Thinking variants of the same size grouped together. Errors are assigned to the earliest stage at which the pipeline fails: skill selection, output parsing/execution, or final reasoning.”}
    \label{fig:error_composition}
\end{figure*}

\paragraph{Error Analysis} Figure ~\ref{fig:error_composition} reveals three notable patterns. First, sub-12B models exhibit predominantly \textbf{skill selection error} on complex benchmarks, up to 98.2\% on InsurBench for gemma-3-270m, consistent with skill routing being a capability that emerges only at larger scales. Second, for models $\geq$30B, selection error is generally near zero and the dominant failures shift to \textbf{parse/execution} and \textbf{reasoning/grounding} errors. For instance, qwen3-30b shows 70.2\% parse error and 8.8\% reasoning error on InsurBench. Third, the ``thinking'' variant displays a mixed profile: it achieves the highest accuracy on IMDB (96.9\%) but incurs higher selection error on FiNER (6.9\%) and InsurBench (49.9\%), as well as elevated parse error on MBPP (27.6\%). This suggests that deliberative reasoning may benefit simpler, well-structured tasks while, in more complex settings, may increase the risk of misselection or instruction-following drift within extended reasoning chains.

\begin{figure}[htbp]
    \centering
    \includegraphics[width=0.48\textwidth]{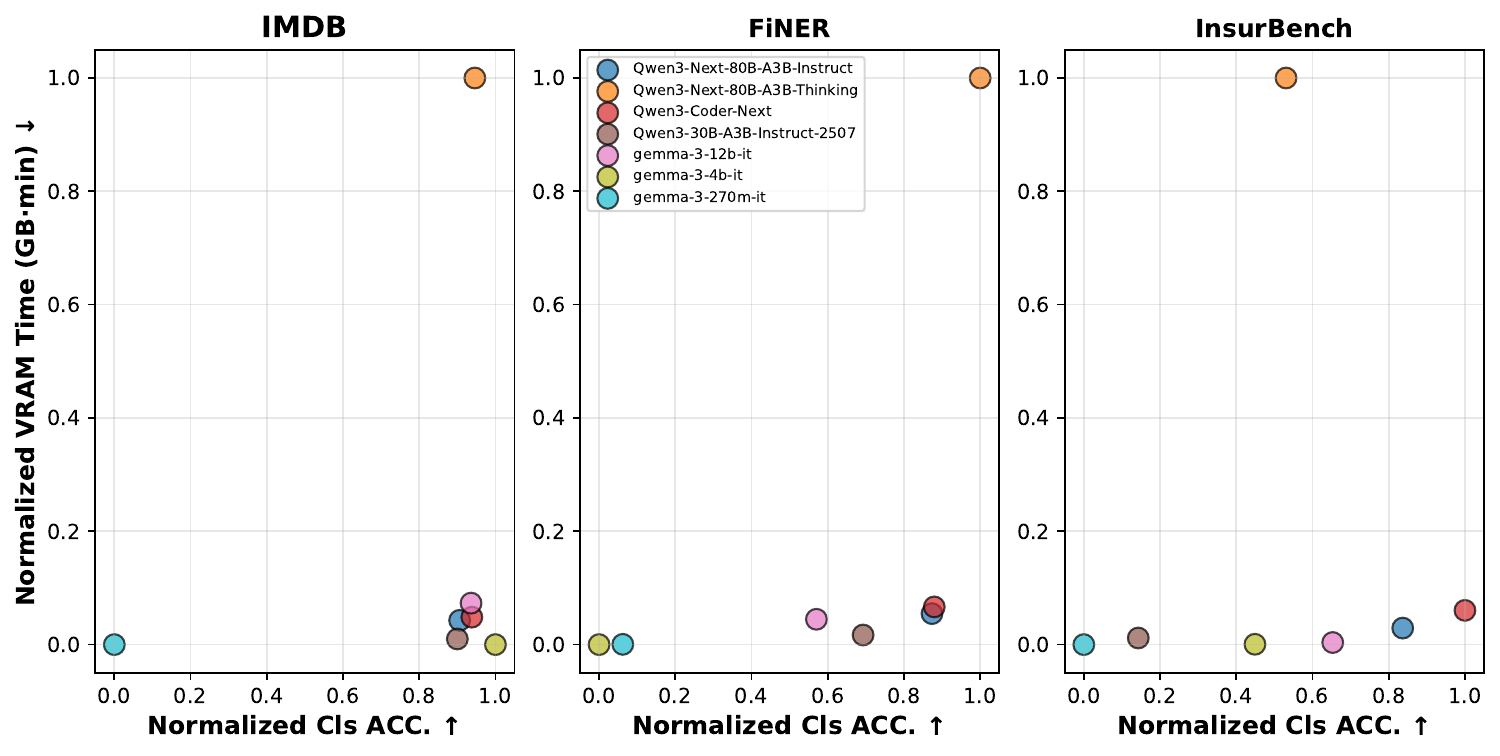}
    \caption{Normalized average VRAM time vs. task performance across different model variants on three datasets. Lower-left region indicates superior performance; both axes are normalized.}
    \label{fig:Vram Efficiency}
\end{figure}

\paragraph{High Cost, Low Return} As shown in Figure \ref{fig:Vram Efficiency}, Qwen3-Next-80B-A3B-Thinking exhibits substantially higher resource consumption than Qwen3-Next-80B-A3B-Instruct under comparable methods. On IMDB, FiNER, and InsurBench, its VRAM Time reaches 24.93, 50.31, and 58.61 GB$\cdot$min (DI), respectively, approximately \(3\times\), \(3.5\times\), and \(2.8\times\) the cost of Instruct (8.47 / 14.52 / 20.69 GB$\cdot$min). Yet accuracy gains are limited: with ASI, Thinking achieves 0.97 Cls ACC on IMDB versus 0.93 for Instruct, but only 0.68 versus 0.63 on FiNER and 0.55 versus 0.63 on InsurBench. On MBPP, Thinking's Pass ACC is 0.71 (ASI) compared with 0.92 for Instruct, despite incurring 251.36 versus 5.34 GB$\cdot$min. These results suggest that the reasoning mechanism generates considerable additional tokens without commensurate improvements in task accuracy, and in some cases underperforms the non-thinking variant, highlighting a structural tension between computational cost and return that warrants careful consideration before deployment.

\begin{figure}[htbp]
    \centering
    \includegraphics[width=0.48\textwidth]{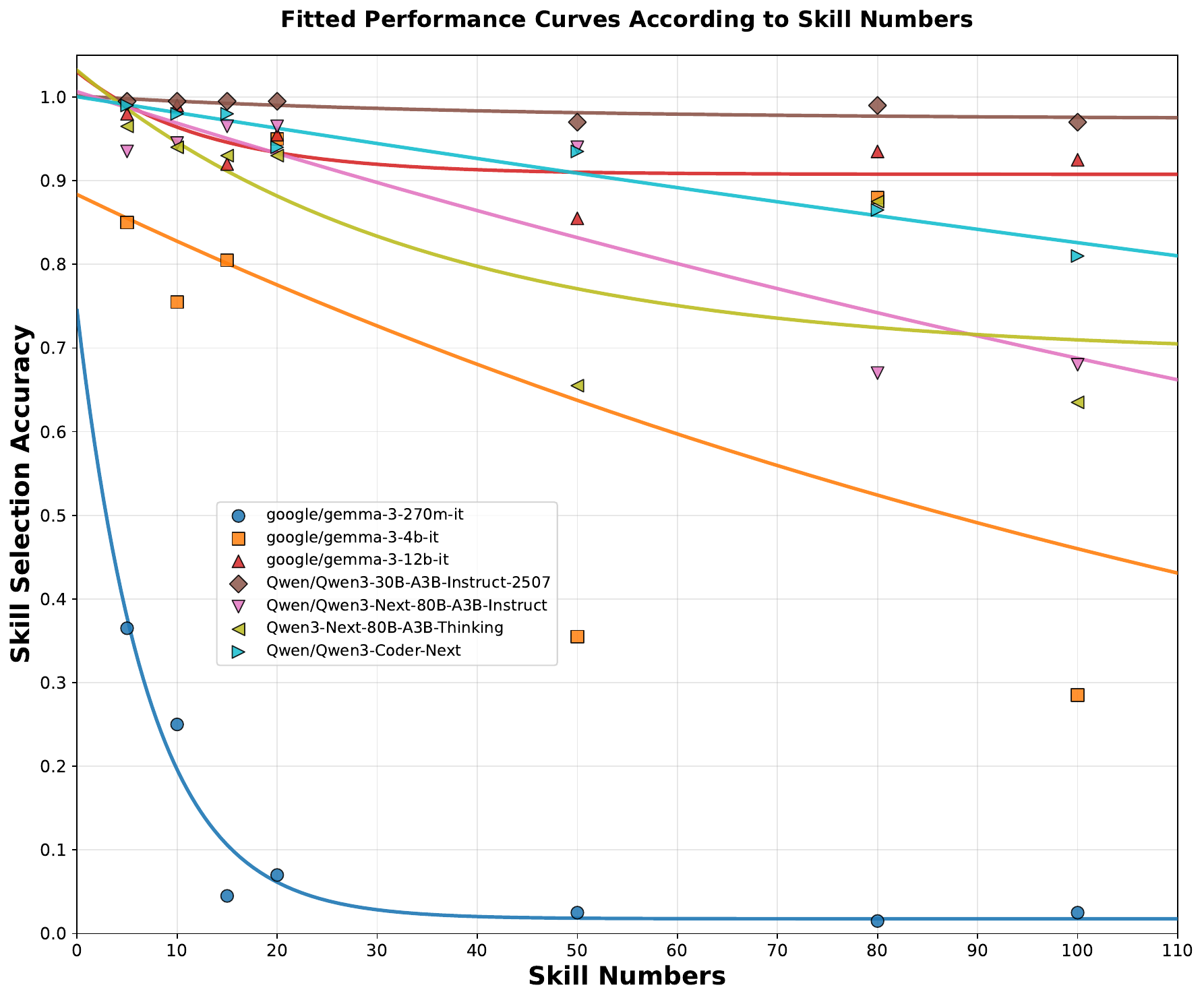}
    \caption{Fitted decay curves (solid lines) and empirical data points (markers) for skill selection accuracy across skill number $N = 5$ to $100$. }
    \label{fig:skill-selection-performance}
\end{figure}

\subsection{Post‑hoc Exploration}

\begin{table}[htbp]
\centering
\begin{adjustbox}{max width=0.50\textwidth}
\begin{tabular}{cccc}
\toprule
\textbf{Model Name} & \textbf{Methods} & \textbf{\begin{tabular}[c]{@{}c@{}}Cls\\ ACC. ↑\end{tabular}} & \textbf{\begin{tabular}[c]{@{}c@{}}Avg VRAM\\ Time (GB·min) ↓\end{tabular}} \\
\midrule
\multirow{2}{*}{Qwen3-Next-80B-A3B-Instruct} & ASI & 0.62 & 5.32 \\
 & ASIH & 0.54 & 10.04 \\
 \midrule
\multirow{2}{*}{Qwen3-30B-A3B-Instruct} & ASI & 0.45 & 2.15 \\
 & ASIH & 0.50 & 2.24 \\
 \midrule
\multirow{2}{*}{gemma-3-12b-it} & ASI & 0.58 & 0.71 \\
 & ASIH & 0.59 & 0.75 \\
 \midrule
\multirow{2}{*}{gemma-3-4b-it} & ASI & 0.53 & 0.15 \\
 & ASIH & 0.66 & 0.15 \\
 \midrule
\multirow{2}{*}{gemma-3-270m-it} & ASI & 0.42 & 0.05 \\
 & ASIH & 0.53 & 0.06 \\
\bottomrule
\end{tabular}
\end{adjustbox}
\caption{Performance of Various Language Models on InsurBench: ASI vs. ASIH (ASI with Chat History)}
\label{tab:table_with_history}
\end{table}

\begin{table}[htbp]
\centering
\begin{adjustbox}{max width=0.45\textwidth}
\begin{tabular}{ccccccc}
\toprule
\textbf{Method} &
\textbf{\begin{tabular}[c]{@{}c@{}}Keyword\\ Name\end{tabular}} &
\textbf{\begin{tabular}[c]{@{}c@{}}Cls\\ ACC. $\uparrow$\end{tabular}} &
\textbf{\begin{tabular}[c]{@{}c@{}}Cls\\ F1 $\uparrow$\end{tabular}} &
\textbf{\begin{tabular}[c]{@{}c@{}}Skill\\ ACC. $\uparrow$\end{tabular}} &
\textbf{\begin{tabular}[c]{@{}c@{}}AVG\\ GT (min) $\downarrow$\end{tabular}} &
\textbf{\begin{tabular}[c]{@{}c@{}}Avg VRAM\\ Time (GB$\cdot$min) $\downarrow$\end{tabular}} \\
\midrule
\multirow{5}{*}{ASI} & Skill        & \textbf{0.62} & 0.60          & 0.92          & 0.03          & 5.32 \\
                     & Capability   & 0.60          & 0.59          & 0.92          & 0.03          & 6.15 \\
                     & Expertise    & 0.61          & \textbf{0.61} & \textbf{0.93} & 0.03          & 6.18 \\
                     & Proficiency  & 0.58          & 0.58          & 0.93          & 0.03          & 5.95 \\
                     & Know-how     & 0.58          & 0.58          & 0.91          & \textbf{0.02} & \textbf{4.30} \\
\midrule
\multirow{5}{*}{FSI} & Skill        & 0.53          & 0.51          & -              & 0.04          & 8.20 \\
                     & Capability   & 0.49          & 0.44          & -              & 0.06          & 10.46 \\
                     & Expertise    & \textbf{0.57} & \textbf{0.54} & -              & 0.06          & 10.74 \\
                     & Proficiency  & 0.51          & 0.46          & -              & 0.06          & 10.49 \\
                     & Know-how     & 0.54          & 0.50          & -              & \textbf{0.04} & \textbf{7.46} \\
\bottomrule
\end{tabular}
\end{adjustbox}
\caption{Performance comparison of different skill synonyms under ASI and FSI frameworks on Qwen3-Next-80B-A3B-Instruct tested on InsurBench. Bold values indicate the best performance within each method group.}
\label{tab:other_skill_performance}
\end{table}

\paragraph{Hitting Paradigm? Small Vs Large.} To further investigate the skill routing limit of different models, models are evaluated under larger skill hubs with exceeding 100 skills. As shown in Figure \ref{fig:skill-selection-performance}, tiny models exhibit a rapid decline in accuracy once the number of candidate skills exceeds $N=10$--$20$. In contrast, models with more than 30B parameters demonstrate strong robustness, maintaining high precision even when selecting from $N=100$ skills. Furthermore, for small- and medium-sized models, a skill library of approximately 10--50 skills appears to provide a reliable operating range for skill routing, a configuration recommendation of personal-agent deployments such as Hermes Agents. Finally, SLMs struggle to capture hierarchical, skill-revealing structures, whereas medium- and large-scale models can more reliably handle nested dependencies within a single \texttt{SKILL.md}.

\paragraph{Does chat history matter?} In Table~\ref{tab:table_with_history}, we evaluate the impact of incorporating conversational history on model performance in InsurBench. To avoid context-window overflow, the dialog before each call is truncated, retaining the system prompt and the most recent 3--4 turns. The results indicate that history benefits are largest for very small models (e.g. gemma-3-4b-it and gemma-3-270m-it), while improvements for medium models are only slight. However, chat history increases VRAM time usage---for Qwen3-Next-80B-A3B-Instruct, the cost increases from 5.32 to 10.04~GB$\cdot$min per item. Therefore, enabling chat-history processing is mainly recommended for lightweight SLMs in agent skills deployments.

\paragraph{Replacing "Skill" with "others"?} We conduct one final experiment to also investigate whether replacing the keyword "Skill" with its synonyms affects the efficiency and accuracy of agent tasks. As shown in Table~\ref{tab:other_skill_performance}, four synonyms were tested. Their impact is shown to be minimal on the performance. Notably, "Expertise" consistently outperformed "Skill" across metrics, suggesting it as a potentially superior alternative. Additionally, "Know-how" demonstrated substantial improvements in GPU memory efficiency, with modest degradation.

\section{Conclusion}

We present a large scale deployment-oriented evaluation of agent skills for open-source models under GPU and data-security constraints, spanning models from 270M to 80B parameters. In particular, we evaluate these dimensions together in a realistic industrial setting. We show that very small models ($\leq$4B) frequently fail at skill selection, whereas mid-sized and larger models ($\geq$30B) consistently benefit from skill augmentation. We further did post-hoc exploration of the usage of chat history, replacing ``skill'' with other words and also the upper bound of skill-hub size. From our results, 30B--80B models offer the most favorable accuracy--cost trade-off for skill-augmented pipelines, though the optimal operating point remains task- and budget-dependent.

\section*{Limitations}

We evaluate a representative set of open-source models from 270M to 80B parameters, including instruct, code, and thinking variants. Due to limited resources, we do not cover all available models. Our findings should therefore be interpreted as characterizing trends across model scales and variants, rather than an exhaustive model comparison. In addition, our experiments rely on existing, human-curated open-source skills with a fixed number of distractor skills. We do not study automatically generated skills, skill evolution over time, or large-scale skill repositories beyond 100 entries. Our results thus primarily speak to the feasibility and cost of using skills under controlled conditions. Our GPU cost metrics (Avg GT and Avg VRAM Time) are measured under a specific inference stack and hardware configuration. While the methodology is portable, absolute costs will vary with quantization, batching, serving framework, and cluster topology. We present these metrics as relative indicators for model selection under constrained budgets.

\clearpage
\bibliography{custom}

\appendix



\section{Details For InsurBench}

InsurBench is a long-context decision-making benchmark tailored to insurance claims scenarios, designed to evaluate the understanding and decision-making capabilities of large language models in real-world business workflows. The core task is: given a complete email thread related to a claim, the model must determine whether the insurance company needs to reply and take action based on the latest information in the thread (take action vs.\ no action). Specifically, the model must (i) localize key emails within threads containing dozens of messages; (ii) distinguish automated notification emails (e.g., system-generated attachment delivery notices, auto-sent messages explicitly stating ``do not reply'') from emails requiring human intervention; (iii) understand implicit business logic in the correspondence; and (iv) output a binary decision on whether the insurer must reply to the thread to continue processing the claim. This task not only tests long-text comprehension but also assesses understanding of insurance workflows, communication protocols, and timeliness requirements.

The email correspondence in InsurBench is sourced from real claim cases at an insurance company during 2025--2026. All emails have been de-identified, with personally identifiable information (PII) such as customer names, policy numbers, and addresses removed or generalized, retaining only business content relevant to claim decision-making. The dataset comprises 300 unique claim cases and over 10,000 emails in total (approximately 33--40 emails per case on average). Thread lengths range from a minimum of 5 emails to a maximum of over 100, with a median of approximately 28 emails per thread, ensuring substantial diversity in context length and complexity.

The labeling task is binary classification. \textbf{Take Action} indicates that the insurance company must reply to the email thread to continue processing the claim. \textbf{No Action} indicates that no reply is needed and the thread can be temporarily archived or await further customer action. Annotation was performed by multiple internal claims adjusters who independently reviewed each complete email thread and applied the binary labels according to the definitions above. The entire annotation process employed one-to-one human review to ensure high label quality and business validity. InsurBench was intentionally designed to reflect the class imbalance characteristics of real-world scenarios while avoiding extreme skew that could distort evaluation: approximately 45\% of cases are labeled Take Action and 55\% are labeled No Action.

\section{Skill Modeling as an MDP}

We provide here a mathematical definition for modeling the progressively disclosed agent skill framework. Based on this definition, we offer an intuitive explanation and conjecture as to why skills may progressively enhance the capabilities of an agent. This discussion is intended for reference only; therefore, it is not included in the main content and is presented solely as supplementary intuition and interpretation. This section provides a conceptual formalization of the agent–skill interaction as a Partially Observable Markov Decision Processes (POMDP: \cite{puterman2014markov}) . 

\subsection{Definition}
An agent skill system can be abstracted as a constrained, information-seeking controller over a partially observed world: at each step, the agent decides whether to (i) \textbf{commit} to a skill, (ii) \textbf{reveal} more information about that skill (or its references), or (iii) \textbf{act} in the environment, trading off task progress against limited attention/context. Considering that the underlying system evolves according to Markov dynamics while the agent cannot directly observe the true (agent--task) state, the agent must maintain a \emph{belief} state that summarizes its interaction history. It aggregates past observations and previously taken actions into a posterior probability distribution over the current latent state. Hence, this setting is naturally modeled as a POMDP  augmented with explicit \emph{information-acquisition} actions (e.g., $\mathrm{reveal}(\rho_k)$) that trade off additional cost for improved observations.
Let each skill $k \in \mathcal{K}$ be represented by the abstract triple $k \in \mathcal{K}, k=(d_k,\;\pi_k,\;\rho_k)$, where $d_k$ is a textual descriptor (e.g., the skill's name and description), $\pi_k$ is a temporally extended intra-skill policy (i.e., an option-level procedure or workflow that induces a sequence of actions), and $\rho_k$ is a reference mechanism (formalizable as a Markov kernel) that can reveal additional skill-relevant context and tools, potentially including pointers to other skills. Define a POMDP, $\mathcal{M}=\langle S,A,O,T,\Omega\rangle$,


\noindent\textbf{1.\ State} $s_t \in S$: The (possibly hidden) task situation: user intent, task stage/progress, and unretrieved environment facts.


\noindent\textbf{2.\ Observation} $o_t \in O$: What the agent can access at time $t$, e.g., the current user message, available history, and any revealed/loaded resources and skills.

\noindent\textbf{3.\ Actions} $a_t \in A$: Skill selection $\mathrm{use}(k)$, paid context acquisition $\mathrm{reveal}(\rho_k)$, skill execution $\mathrm{execute}(\pi_k)$, and environment/tool calls (plus internal planning/inference).

\noindent\textbf{4.\ Belief} $b_t = P(s_t \mid o_{\le t}, a_{<t})$: The agent's posterior over $s_t$ given observations so far and past actions (its uncertainty state).



\noindent\textbf{5.\ Transition} $T$ (Markov): How the latent state changes after acting (progress, context/skill availability, and inferred intent updates):
$P(s_{t+1}\mid s_t,a_t)=T(s_{t+1}\mid s_t,a_t)$.


\noindent\textbf{6.\ Observation model} $\Omega$: How the next observation is produced from the updated context and action outcomes (e.g., reveal results, tool outputs); the new observation then drives a belief update for the next decision.

\subsection{Intuitive interpretation}
Model capability (typically increasing with model size via scaling laws) determines how accurately an agent can approximate the optimal belief from observations (user messages, revealed skill content, and tool outputs) given the observation history. When the agent is highly uncertain, as reflected by a diffuse belief state $b_t$, it is often worthwhile to spend additional time and cost to \emph{reveal} relevant skill context before committing to execution, since the expected value of information can outweigh the reveal cost. When the agent is already nearly certain (i.e., its belief is concentrated after acquiring sufficient evidence), directly executing the appropriate workflow description is typically more cost-effective than further disclosure. This progressive-disclosure behavior in agent skill aligns with the classic result that, for finite-horizon POMDPs, the optimal value function over the belief space is piecewise-linear and convex \cite{KAELBLING199899}: different regions of belief correspond to different optimal contingent plans. Consequently, within an agent framework, one can view progressive disclosure as an instance of approximately optimal control with \emph{information actions}: reveal only what is necessary to increase expected utility (by reducing uncertainty) before executing the intra-skill policy, rather than exhaustively loading all context upfront.

\subsection{Proposition: Skill Use Improves the Value Lower Bound}

We formalize the intuition that, \emph{conditioned on}  
(A1) correct skill selection and  
(A2) skill correctness and specialization,  
invoking a skill cannot decrease the agent's achievable value relative to acting without the skill.

The proof proceeds in three steps:
\begin{enumerate}[label=(\roman*)]
    \item Cast the belief MDP induced by $\mathcal{M}$.
    \item Show that $\mathrm{reveal}(\rho_k)$ is a Blackwell-informative refinement of the belief.
    \item Show that the composite option $\mathrm{use}(k) + \mathrm{execute}(\pi_k)$ is subgoal-optimal and dominates the primitive-action baseline under (A1)--(A2).
\end{enumerate}

\subsubsection{Setup: the belief-MDP}

Since $\mathcal{M}=\langle S,A,O,T,\Omega\rangle$ is a POMDP, it is standard \citep{puterman2014markov} that $\mathcal{M}$ induces an equivalent fully observed \emph{belief-MDP} $\mathcal{M}^b = \langle B, A, T^b, R^b\rangle$ over the belief simplex $B=\Delta(S)$, with belief update
\begin{equation}
\label{eq:belief-update}
\begin{aligned}
b_{t+1}(s')
&= \frac{\Omega(o_{t+1}\mid s',a_t)\,\sum_{s} T(s'\mid s,a_t)\,b_t(s)}{\Pr(o_{t+1}\mid b_t,a_t)}.
\end{aligned}
\end{equation}

and value function satisfying the Bellman optimality equation
\begin{equation}
\label{eq:belief-bellman}
\begin{aligned}
V^\ast(b)
&= \max_{a\in A}\Big[
R^b(b,a) + \gamma \sum_{o} \Pr(o\mid b,a)\\
&\qquad\qquad\cdot V^\ast(\mathrm{update}(b,a,o))\Big].
\end{aligned}
\tag{1}
\end{equation}
$V^\ast$ is well known to be piecewise-linear and convex in $b$ for finite horizon \cite{10.1287/opre.21.5.1071}; we only need convexity of $V^\ast(\cdot)$ in $b$, which holds regardless of horizon truncation.

We assume rewards are bounded, $R^b\in[0,R_{\max}]$, and discounting $\gamma\in(0,1]$ (or a finite horizon $H$), so all values below are finite.

\subsubsection{Assumptions}

\textbf{(A1) Correct skill selection.} There exists a (possibly latent) ground-truth "compatible" skill set $\mathcal{K}^\star(s)\subseteq\mathcal{K}$ for state $s$, and the selection event
\begin{equation}
\label{eq:Esel}
\begin{aligned}
\mathcal{E}_{\text{sel}}
&= \{\, \mathrm{use}(k) \text{ is chosen at } t \\
&\qquad\text{and } k\in\mathcal{K}^\star(s_t) \,\}.
\end{aligned}
\tag{A1}
\end{equation}
holds, i.e. the agent commits only to skills whose applicability precondition is actually satisfied by the latent state. Formally, we condition the whole analysis on $\mathcal{E}_{\text{sel}}$.

\paragraph{(A2) Skill correctness / specialization.} Conditioned on $k \in \mathcal{K}^\star(s_t)$:
\begin{itemize}
\item[(A2a)] (\emph{Reveal is informative and non-corrupting}) $\rho_k$ is a Markov kernel from $s$ to a refined observation $o'$ such that $o'$ is a \emph{sufficient statistic refinement} of the default observation $o$ in the Blackwell sense: there is no information destruction, i.e. $o = f(o')$ for some (possibly stochastic) coarsening $f$, and $\rho_k$ does not inject label noise (no adversarial/incorrect content).
\item[(A2b)] (\emph{Intra-skill policy is at least as good as the best primitive alternative on its own domain}) For every belief $b$ supported on $\{s: k\in\mathcal{K}^\star(s)\}$,
\begin{equation}
\label{eq:A2b}
\begin{aligned}
Q^{\pi_k}(b)
&:= \mathbb{E}_{\substack{\text{start belief } b;\\ \text{follow } \pi_k \text{ until termination;}\\ \text{then follow } \pi^\ast}}
\Bigg[\sum_{t} \gamma^t r_t\Bigg] \\
&\ge V^{\ast}_{\text{no-skill}}(b),
\end{aligned}
\tag{A2b}
\end{equation}
where $V^\ast_{\text{no-skill}}$ is the optimal value attainable using only primitive actions (no $\mathrm{use}/\mathrm{reveal}/\mathrm{execute}$), and both are evaluated net of any invocation cost $c_k\ge 0$.
\end{itemize}

(A2a) says the skill's reference/content is \emph{correct} (truthful, non-misleading); (A2b) says it is \emph{specialized}, i.e. professionally engineered to be a good — not necessarily unique — procedure for the sub-task it targets.

\subsubsection{Step 1 — Reveal weakly improves value (Blackwell monotonicity)}

\textbf{Lemma 1 (Value of an informative reveal).} Let $b_t$ be the current belief and let $b_t^{o'}$ denote the belief after taking $a_t=\mathrm{reveal}(\rho_k)$ and observing $o'\sim\rho_k(\cdot\mid s_t)$, with $c_{\text{rev}}\ge 0$ the cost of revealing. Under (A2a),
\begin{equation}
\label{eq:lemma1-ineq}
\begin{aligned}
\mathbb{E}_{o'}\big[V^\ast(b_t^{o'})\big] - c_{\text{rev}}
&\ge V^\ast(b_t) - c_{\text{rev}}^{\min}.
\end{aligned}
\tag{L1}
\end{equation}
whenever $c_{\text{rev}}$ is dominated by the expected information gain, i.e. skills are only revealed when

\begin{equation}
\label{eq:rational-reveal}
\begin{aligned}
c_{\text{rev}}
&\le \mathbb{E}_{o'}\big[V^\ast(b_t^{o'})\big] - V^\ast(b_t).
\end{aligned}
\tag{2}
\end{equation}

\emph{Proof.} $V^\ast(\cdot)$ is convex on $B$ (piecewise-linear-convex belief value, standard POMDP result). By Jensen's inequality applied to the martingale property of Bayesian belief updates ($\mathbb{E}_{o'}[b_t^{o'}] = b_t$ under the true generative model),
\begin{equation}
\label{eq:blackwell-jensen}
\begin{aligned}
\mathbb{E}_{o'}\big[V^\ast(b_t^{o'})\big]
&\ge V^\ast\big(\mathbb{E}_{o'}[b_t^{o'}]\big) \\
&= V^\ast(b_t).
\end{aligned}
\tag{3}
\end{equation}
This is exactly Blackwell's theorem specialized to POMDP belief states: any signal $o'$ that is a mean-preserving refinement of the current belief cannot decrease the expected optimal value, because the decision-maker can always ignore $o'$ and act as if $b_t$ were unchanged (weak improvement), and strictly benefits whenever $V^\ast$ is strictly convex at $b_t$ in the direction induced by $\rho_k$ (strict improvement) \citep{blackwell1953equivalent}. (A2a) is exactly the assumption ruling out the degenerate/adversarial case where $\rho_k$ could \emph{decrease} the informativeness of $o$ (a corrupting or misleading reveal, which would break the martingale/refinement property and could lower $V^\ast$ in expectation). Subtracting the cost $c_{\text{rev}}$ and imposing the rational-invocation condition (2) gives the claim. Lemma 1 formalizes ``revealing more about a \emph{correct} skill cannot hurt, and typically helps'' — it is the information-acquisition half of the framework.

\subsubsection{Step 2 — Using a correctly selected, specialized skill weakly dominates the no-skill policy}

We model $(\mathrm{use}(k),\mathrm{execute}(\pi_k))$ as inducing a temporally extended \emph{option} $o_k=\langle \mathcal{I}_k,\pi_k,\beta_k\rangle$ in the sense of \citet{SUTTON1999181}, with initiation set $\mathcal{I}_k = \{s: k \in \mathcal{K}^\star(s)\}$ (guaranteed nonempty and correctly entered under (A1)), internal policy $\pi_k$, and termination condition $\beta_k$. Let $\Pi_{\text{prim}}$ denote the space of policies over primitive actions only, and $\Pi_{\text{skill}} = \Pi_{\text{prim}} \cup \{o_k : k\in\mathcal{K}\}$ the augmented policy space obtained by adding the skill-option.

\textbf{Lemma 2 (Skill-augmented value dominance).} Conditioned on $\mathcal{E}_{\text{sel}}$ (A1) and (A2b), for every belief $b\in \mathcal{I}_k$ (lifted to belief space),
\begin{equation}
\label{eq:lemma2-dominance}
V^\ast_{\Pi_{\text{skill}}}(b) \;\ge\; V^\ast_{\Pi_{\text{prim}}}(b).
\tag{4}
\end{equation}

\emph{Proof.} Because $\Pi_{\text{prim}} \subseteq \Pi_{\text{skill}}$ (adding an option to the action set is a strict superset of available policies — the agent always retains the ability to ignore the skill and fall back to primitive actions), we trivially have $V^\ast_{\Pi_{\text{skill}}}(b) \ge V^\ast_{\Pi_{\text{prim}}}(b)$ by monotonicity of the max operator over a superset of policies. This is the general "options cannot hurt in the limit of optimal planning" fact \citep{SUTTON1999181}. What (A2b) adds is a \emph{quantitative, non-asymptotic} guarantee: it directly asserts $Q^{\pi_k}(b) \ge V^\ast_{\text{no-skill}}(b)$, i.e. the specific policy $\pi_k$ realizes the sup, so the inequality (4) is not just non-strict in principle but is achieved by the concrete, deployed skill rather than requiring re-optimization over all options. Formally, by the Bellman equation for the option-augmented MDP,
\begin{equation}
\label{eq:option-bellman-chain}
\begin{aligned}
V^\ast_{\Pi_{\text{skill}}}(b)
&= \max\Big\{ V^\ast_{\Pi_{\text{prim}}}(b),\; Q^{\pi_k}(b) - c_k \Big\} \\
&\ge Q^{\pi_k}(b) - c_k \\
&\ge V^\ast_{\text{no-skill}}(b).
\end{aligned}
\end{equation}
where the last step is (A2b), and combining with the trivial bound gives (4).

\subsubsection{Step 3 — Composed lower bound}

\textbf{Proposition (Progressive disclosure improves the achievable value lower bound).} Condition on $\mathcal{E}_{\text{sel}}$ (correct selection, A1) and on the skill being correct and specialized (A2a-A2b). Let $V^\ast_{\mathcal{M}}(b_0)$ denote the optimal value of the full skill-augmented POMDP from initial belief $b_0$, and $V^\ast_{\text{no-skill}}(b_0)$ the optimal value with no skill mechanism (no $\mathrm{use}/\mathrm{reveal}/\mathrm{execute}$ actions available, i.e. the agent must solve the task from primitive actions and raw observations alone). Then

\begin{equation}
\label{eq:composed-lb}
\begin{aligned}
V^\ast_{\mathcal{M}}(b_0)
&\ge V^\ast_{\text{no-skill}}(b_0) - (c_{\text{rev}} + c_k) \\
&\qquad\qquad + \Delta_{\text{info}}(b_0,\rho_k).
\end{aligned}
\tag{5}
\end{equation}

\begin{equation}
\label{eq:delta-info-def}
\begin{aligned}
\Delta_{\text{info}}(b_0,\rho_k)
&:= \mathbb{E}_{o'}\big[V^\ast(b_0^{o'})\big] - V^\ast(b_0) \\
&\ge 0.
\end{aligned}
\end{equation}

In particular, whenever the rational-invocation condition (2) holds (the agent only pays for reveals/skills whose expected benefit covers their cost), the right-hand side is $\ge V^\ast_{\text{no-skill}}(b_0)$, hence
\begin{equation}
\label{eq:main-dominance}
V^\ast_{\mathcal{M}}(b_0) \;\ge\; V^\ast_{\text{no-skill}}(b_0).
\tag{6}
\end{equation}
Equality holds only in the degenerate case where the skill is uninformative relative to $o$ ($\rho_k$ a trivial/constant kernel) and $\pi_k$ exactly replicates the optimal primitive policy; strict inequality $V^\ast_{\mathcal{M}}(b_0) > V^\ast_{\text{no-skill}}(b_0)$ holds whenever either $V^\ast(\cdot)$ is strictly convex along the direction of the belief refinement induced by $\rho_k$, or $\pi_k$ strictly dominates the best primitive policy on $\mathcal{I}_k$.

\emph{Proof.} Combine Lemma 1 (applied at the reveal step reached along the optimal trajectory, giving the $\Delta_{\text{info}} \ge 0$ term net of $c_{\text{rev}}$) with Lemma 2 (applied at the use/execute step, giving the $Q^{\pi_k}-c_k \ge V^\ast_{\text{no-skill}}$ term), both conditioned on $\mathcal{E}_{\text{sel}} \cap \{k \text{ correct, specialized}\}$. Since these are two sequential decision points on the same trajectory and $V^\ast_{\mathcal{M}}$ is obtained by \emph{jointly} optimizing over reveal-then-use-then-execute vs. any primitive alternative (the skill actions are additional options in the same max in the Bellman equation, cf. Eq. 1), the joint optimum cannot be worse than either the reveal-only or use-only bound, and in particular cannot be worse than not using the skill mechanism at all — which is exactly $V^\ast_{\text{no-skill}}(b_0)$. Formally, unrolling Eq. (1) at $b_0$ with the skill branch explicit:
\begin{equation}
\label{eq:unroll-bellman-b0}
\begin{aligned}
V^\ast_{\mathcal{M}}(b_0)
&= \max\Big\{
V^\ast_{\text{no-skill}}(b_0),\\
&\qquad -c_{\text{rev}} + \mathbb{E}_{o'}\big[-c_k + Q^{\pi_k}(b_0^{o'})\big]
\Big\}.
\end{aligned}
\end{equation}
and the second term is $\ge V^\ast_{\text{no-skill}}(b_0)$ term-by-term by Lemmas 1-2 whenever (2) holds, so the outer $\max$ equals (is lower-bounded by) that branch, yielding (5)-(6).

\subsubsection{Interpretation}

Equation (6) is the formal counterpart of the paper's intuition: \emph{skills cannot hurt, and generically help, but only once selection is correct and the skill itself is trustworthy and well-designed for its niche.} Both conditioning assumptions are load-bearing and cannot be dropped:

\begin{itemize}
\item If (A1) fails — i.e. the agent invokes a skill outside its true applicability region $\mathcal{K}^\star(s)$ — then $\mathcal{I}_k \not\ni s_t$ and (A2b)'s guarantee is vacuous; $\pi_k$ may actively steer the agent away from the optimal trajectory (negative transfer), so no lower bound of the form (6) can be derived, and empirically this is where progressive disclosure architectures must invest in a reliable router/selector.
\item If (A2a) fails (the skill's $\rho_k$ is stale, wrong, or misleading), Lemma 1's martingale argument breaks: $o'$ is no longer a Blackwell-refinement of $o$, so $\mathbb{E}_{o'}[V^\ast(b_t^{o'})]$ can fall below $V^\ast(b_t)$ (misinformation can be worse than no information).
\item If (A2b) fails (the skill is technically applicable but poorly engineered — e.g. an inefficient or buggy workflow), $Q^{\pi_k}$ can fall below $V^\ast_{\text{no-skill}}$, and the dominance in Lemma 2 reverses.
\end{itemize}

This is why the statement is phrased as a \emph{conditional} guarantee (a conjecture/intuition, not an unconditional theorem about all skill systems): it isolates \emph{skill quality and routing accuracy} as the two necessary and sufficient levers, matching the ablations one would want in the empirical sections of the paper (varying router accuracy and skill quality independently and checking that measured performance tracks the two right-hand-side terms in (5)).

\section{Additional Experimental Settings}

In all experiments, the models' default decoding configurations (e.g., temperature, top-\(p\), and top-\(k\)) were used to avoid confounding effects from additional hyperparameter tuning. All methods were implemented using the LangChain agent framework to ensure a consistent agentic workflow across settings. For inference, vLLM was used as the serving engine, and the context length was fixed to \(10240\) tokens.

To manage conversational history under the context-window constraint, a deterministic message-trimming policy was applied. Specifically, the first message (typically the system prompt) was always retained, and only the most recent \(3\) or \(4\) messages were kept depending on the parity of the total message count. Core instructions were preserved while the prompt length was bounded, thereby reducing truncation risk and stabilizing inference cost in longer interactions. For formatting and metadata extraction from \texttt{SKILL.md}, the publicly available codebase is primarily used.\footnote{\url{https://github.com/agentskills/agentskills/tree/main/skills-ref}}

In our experiments, we report reserved VRAM rather than allocated VRAM to reflect realistic deployment constraints. When using inference engines such as vLLM or SGLang, all memory reserved by the engine is unavailable to other processes and thus represents the true operational footprint. All reported latencies exclude model-loading time and are measured as the elapsed time between issuing an API request and receiving the completed response.

\section{Additional results}

\begin{table*}[t]
\centering
\caption{Error composition by model and benchmark. Each row sums to 100\% and is decomposed into retrieval error, parsing/execution error, reasoning/grounding error, and correct predictions.}
\label{tab:error_composition}
\small
\begin{adjustbox}{width=\textwidth,center}
\begin{tabular}{llrrrr}
\toprule
Benchmark & Model & Retrieval. Err. & Parse Err. & Reason. Err. & Correct \\
\midrule
\multirow{7}{*}{IMDB} 
 & gemma-3-270m-it & 48.5 & 3.9 & 0.0 & 47.5 \\
 & gemma-3-4b-it & 0.0 & 3.3 & 7.1 & 89.6 \\
 & gemma-3-12b-it & 0.0 & 5.1 & 3.4 & 91.5 \\
 & Qwen3-30B-Instruct & 0.0 & 0.0 & 5.1 & 94.9 \\
 & Qwen3-Coder-Next & 0.0 & 7.1 & 3.8 & 89.1 \\
 & Qwen3-Next-80B-A3B-Instruct & 0.6 & 0.0 & 6.9 & 92.5 \\
 & Qwen3-Next-80B-A3B-Thinking & 0.0 & 0.0 & 3.1 & 96.9 \\ \midrule
\midrule
\multirow{7}{*}{FiNER} 
 & gemma-3-270m-it & 96.3 & 1.5 & 0.1 & 2.1 \\
 & gemma-3-4b-it & 0.1 & 3.3 & 76.1 & 20.5 \\
 & gemma-3-12b-it & 0.0 & 1.5 & 52.0 & 46.5 \\
 & Qwen3-30B-Instruct & 0.4 & 37.0 & 26.3 & 36.3 \\
 & Qwen3-Coder-Next & 0.8 & 11.9 & 29.9 & 57.4 \\
 & Qwen3-Next-80B-A3B-Instruct & 0.6 & 1.5 & 35.1 & 62.9 \\
 & Qwen3-Next-80B-A3B-Thinking & 6.9 & 0.0 & 25.4 & 67.7 \\ \midrule
\midrule
\multirow{8}{*}{InsurBench} 
& gemma-3-270m-it & 98.2 & 0.0 & 0.0 & 1.8 \\
& gemma-3-4b-it & 92.5 & 0.0 & 0.0 & 7.5 \\
& gemma-3-12b-it & 44.4 & 0.0 & 0.0 & 55.6 \\
& Qwen3-30B-Instruct & 0.0 & 70.2 & 8.8 & 21.1 \\
& Qwen3-Coder-Next & 0.6 & 31.6 & 18.9 & 48.9 \\
& Qwen3-Next-80B-A3B-Instruct & 0.3 & 19.8 & 17.0 & 62.9 \\
& Qwen3-Next-80B-A3B-Thinking & 49.9 & 0.0 & 0.0 & 50.1 \\  \midrule
\midrule
\multirow{7}{*}{MBPP} & gemma-3-270m-it & 10.0 & 86.4 & 3.6 & 0.0 \\
 & gemma-3-4b-it & 0.0 & 34.5 & 0.8 & 64.7 \\
 & gemma-3-12b-it & 0.0 & 18.4 & 0.0 & 81.6 \\
 & Qwen3-30B-Instruct & 0.0 & 11.4 & 0.0 & 88.6 \\
 & Qwen3-Coder-Next & 0.0 & 5.8 & 0.0 & 94.2 \\
 & Qwen3-Next-80B-A3B-Instruct & 0.0 & 7.6 & 0.0 & 92.4 \\
 & Qwen3-Next-80B-A3B-Thinking & 0.0 & 27.6 & 1.0 & 71.4 \\
 
\bottomrule
\end{tabular}
\end{adjustbox}
\end{table*}

\paragraph{Agent Skill Calling in the Wild}

\begin{figure}[htbp]
    \centering
    \includegraphics[width=0.38\textwidth]{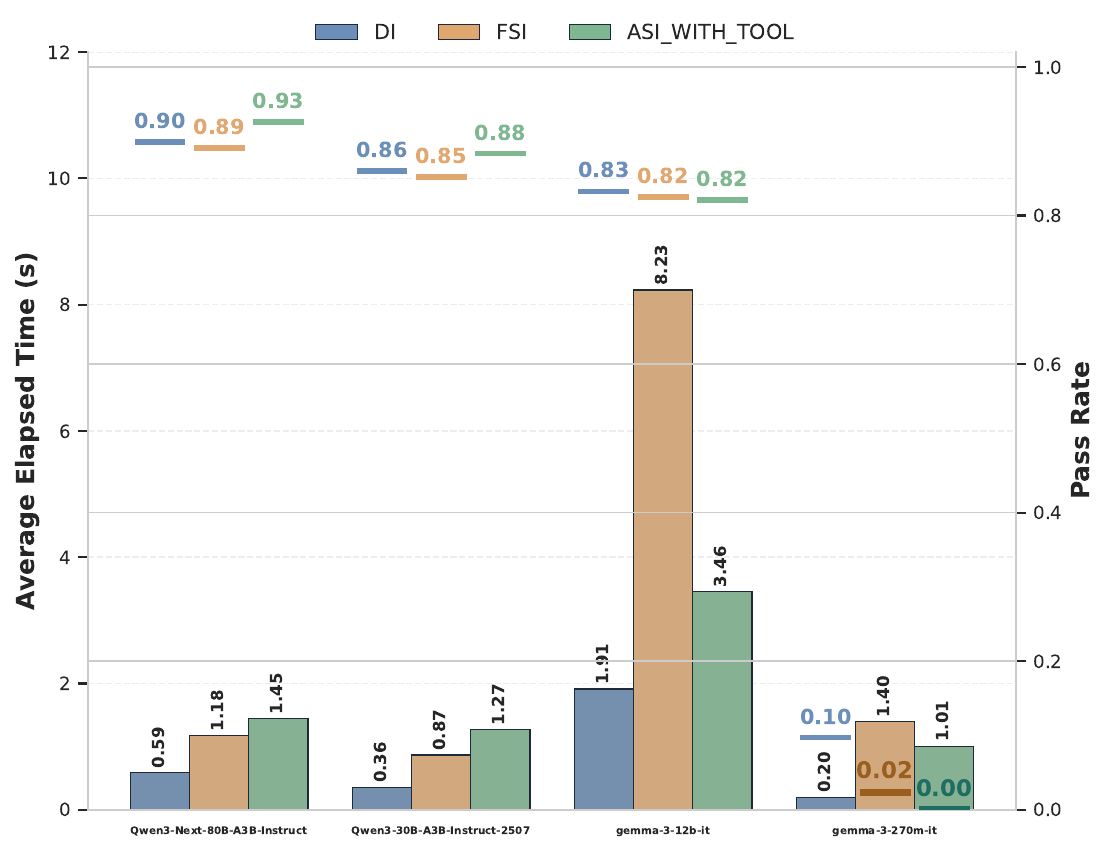}
    \caption{Coding agents performance: bars show average runtime, lines show the average pass rate.}
    \label{fig:skill-mbpp-plot}
\end{figure}

The main experiments above focus on assessing the agent in selecting the appropriate skills and perform complex reasoning based on them, with the design intentionally isolating the effect of skill usage. As for MBPP, which is a ReAct-style environment within a tool calling setting of an integrated Python interpreter, the agent is required to choose among five distractor skills, invoke the selected skill, and following its description for calling the tool to get feedback until reaching a final solution. We report additionally the pass rate and the average execution time in Figure~\ref{fig:skill-mbpp-plot}. The results are consistent with those observed in the no-tool setting: very small models (e.g., 270M) struggle to benefit from the agent skill, whereas larger models (e.g., 30B and 80B) show clear performance gains under agent skill settings.

\section{Error Diagnostic}

\subsection{FiNER}

\begin{responsebox}{Example A --- Input: }
Excluding reimbursable revenues of \$26.3 million, ENL constituted approximately 
17.7 percent of our total consolidated revenues and approximately 46.6 percent 
of our International \& Alaska Drilling segment revenues.
Target numeric entity to tag: \texttt{26.3}
\end{responsebox}

\begin{responsebox}{Model output:  -- gemma-3-270m-it }
  The Excluding Revenues of \$ 26.3 million were 
  approximately 17.7\% of our total consolidated revenues, 
  and approximately 46.6\% of our International \& Alaska 
  Drilling segment revenues. For the nine months ended ...

  Correct label:RevenueFromContractWithCustomer\\ExcludingAssessedTax
\end{responsebox}

\paragraph{Why the model failed.} The sentence mentions the figure \texttt{26.3} in the context of 
``excluding reimbursable revenues'', which corresponds to the specific XBRL 
concept ``Revenue From Contract With Customer Excluding Assessed Tax''. Instead, the model effectively grounds the number on the generic concept  \texttt{Revenues}. This failure reflects both a skill-retrieval issue and also a grounding error: the smallest model frequently fails to load the most appropriate tagging skill and 
relies on coarse financial concepts. The output is well-formed but maps the numeric entity to the wrong financial concept.

\subsection{InsurBench}

\paragraph{Why it is difficult but useful?} InsurBench is a particularly informative hard case because it is
built from real historical corporate email threads and requires a
narrow procedural judgment: given the complete thread, the system must
decide whether the email history especially the latest email creates a present
obligation for the target insurer to act.

The task deliberately combines misleading operational cues with
non-action outcomes. Phrases such as \textbf{``please find attached''}
normally suggest that a document should be reviewed or processed, and
an alternative contact address may appear to invite a reply or
follow-up. Yet many cases are merely automatic delivery notices,
containing limiting cues such as \textbf{``sent automatically''},
\textbf{``do not reply''}, and \textbf{``if needed, contact us''}.
Thus, the benchmark contains \textbf{action-looking non-action cases}:
multilingual, attachment-bearing emails resemble requests, while their
actual function is only to notify or deliver information.

The full history adds a second difficulty. It may include genuine
claims, document requests, payment discussions, liability decisions,
and reminders. The model must use that context to interpret the newest
message, but must not let real historical requests \textbf{leak} into
the current decision. Long, multi-turn threads can therefore create
``lost in the middle'' effects, especially for small and medium-sized
models, which may over-weight salient customer requests in the history.

Finally, the task requires role-sensitive reasoning. The model must
identify the sender and recipient of the latest message, resolve
references such as \textit{nous}, \textit{vous}, and \textit{votre
assuré}, determine who is asked to act, and verify that this party is
the target insurer. In threads involving forwarding, automatic copies,
brokers, counterpart insurers, claimants, and internal teams, merely
detecting a request is not enough: the request must be directed to the
target insurer and originate from the latest message itself.
\subsubsection{Example A}

\begin{responsebox}{Example A --- Input: }
TASK: decide YES/NO --- must the insurer act NOW because of the
LATEST email only?

LATEST EMAIL:

\textbf{French:}
Veuillez trouver, en annexe, un document qui vous est destiné.
\textbf{Nous vous prions de ne pas répondre au présent courriel,
qui vous a été adressé de manière automatique.}
\ldots

\textbf{German:}
Hiermit erhalten Sie ein Dokument zu Ihrer Information.
\textbf{Bitte antworten Sie nicht auf diese automatisch erstellte
Email.}
\ldots

\textbf{English:}
Please find attached a document for your information.
\textbf{Do not reply to this email, which has been sent
automatically.}
\ldots
\end{responsebox}

\begin{responsebox}{Example A: Output qwen3-30b-a3b-instruct }
``\textbf{Yes}, the insurance company must take action now. The email
indicates an intention to indemnify the insured party and requests a
\textbf{property-damage statement} with supporting documentation.
This clearly requires the insurer to process the claim \ldots''
\end{responsebox}

\paragraph{Why the model failed.} The correct label is \textbf{NO}: the latest message is explicitly an
automatically generated delivery notification. Its strongest pragmatic
signals are repeated in three languages: \textbf{``do not reply''} and
\textbf{``sent automatically''}. It mentions an attachment but does
not, by itself, request a response or impose an immediate action
obligation on the insurer. The model instead predicts \textbf{YES} by treating claim-related
language---an attachment, indemnification context, and a request for supporting documentation---as if it were a current request in the
latest message. This illustrates a central InsurBench difficulty:
the message combines operationally suggestive surface cues with
explicit non-action cues, and the model must distinguish the latest
auto-notification from potentially actionable content elsewhere in the email thread.

\subsubsection{Example B}

\begin{responsebox}{Example B --- Input: }
TASK: decide YES/NO --- must the insurer act NOW because of the
LATEST email only?

LATEST EMAIL:

From: \texttt{[SENDER ORGANISATION]}
To: \texttt{[INSURER]}
Subject: \textbf{[CASE-ID] --- 4\textsuperscript{th} REMINDER}

Warning: Replies to this message will go to \texttt{[redacted]}.
\ldots

Bonjour,

\textbf{Merci de prendre connaissance du document joint au présent
message.}

Cordialement,

--- RELEVANT THREAD CONTEXT ---

Earlier correspondence concerning the same claim states:

\textbf{``We have sent you three previous letters regarding our
claim. To date, we have not received a response from you.''}

\textbf{``Please process our claim as soon as possible. Otherwise, we may be compelled to initiate a non-amicable recovery procedure.''}

\ldots
\end{responsebox}

\begin{responsebox}{Example B: Output qwen3-30b-a3b-instruct}

``The most recent message is a follow-up email. Although its body
only says \textbf{`please review the attached document'}, the subject
line explicitly labels it as a \textbf{fourth reminder}. Earlier
messages state that no response has been received and request action
within a short timeframe, with possible escalation.

This is a reminder of previous unanswered correspondence. The insurer
must take action because the sender is following up on an unresolved
claim and has indicated potential recovery proceedings \ldots''
\end{responsebox}

\paragraph{Why the model failed.} The model’s reasoning identifies the decisive evidence correctly:
the latest message is part of an escalated reminder chain, with
\textbf{three prior unanswered requests} and a threat of
\textbf{non-amicable recovery}. The correct label is therefore
\textbf{YES}. However, the model never emits an explicit final
\textbf{YES} or \textbf{NO} decision because it forgets the initial instructions. This sample also illustrates why InsurBench is intrinsically
difficult: the current email body alone is merely
\textbf{``please review the attached document''}; its actionability is
encoded jointly in the \textbf{subject line} (“4th reminder”), the
attachment relation, and the immediately preceding correspondence. The model must use the thread to identify the pragmatic function of the latest message, while not mechanically treating every historical request as a newly issued request.

\subsection{IMDB}

\begin{responsebox}{Example B --- Input}
TASK: classify the review sentiment as \texttt{positive} or \texttt{negative}.

REVIEW:
Wirey's journey through the final days of bachelorhood, liberally sprinkled with 
flashbacks to a sexually active and diverse childhood. It's definitely not a 
feel-good romance movie. It is a romance movie, but one without \ldots
\end{responsebox}

\begin{responsebox}{Example B --- Output: Gemma-3-12B}
Okay, I need to classify the sentiment of this movie review.
I will use the "Sentiment Analysis" skill.
**Sentiment Analysis Skill**
**Description:** This skill analyzes text to determine its sentiment 
(positive, negative, or neutral). It ...
Predicted sentiment: \texttt{negative}
\end{responsebox}

\paragraph{Why the model failed.}
The correct label is again \textbf{positive}. The model focuses on the phrase ``not a feel-good romance movie'' and interprets it as evidence of negative sentiment. It fails to recognize that the reviewer is distinguishing this film from conventional romances while still evaluating it positively. This is a typical grounding failure for mid-sized models on IMDB: the sentiment skill is correctly selected, but the global polarity is misclassified.

\subsection{MBPP}

\begin{responsebox}{Example A --- Input}
TASK: write a Python function that passes all unit tests.

PROMPT:
Write a function to find the tuple intersection of elements in the given tuple 
list irrespective of their order.
\end{responsebox}

\begin{responsebox}{Example A --- Output: Gemma-3-270M}
def find tuple intersection(tuple list):
    """ Finds the tuple intersection of a list of tuples.
    Args:
        tuple list: A list of tuples.
    Returns:
        A tuple containing the intersection of the first and last elements
        of the list. Returns an empty ...
\end{responsebox}

\paragraph{Why the model failed.}
The correct behavior requires a valid Python function that computes the 
intersection of tuples in a list and passes all MBPP unit tests. The generated 
code contains multiple errors: the function name \texttt{find tuple intersection} 
and parameter name \texttt{tuple list} contain spaces and are therefore invalid 
identifiers; the docstring describes an incorrect specification (intersection of 
``first and last elements'' rather than all tuples); and the function body is 
incomplete. These issues cause a \texttt{NameError} or \texttt{SyntaxError} at 
runtime. This illustrates a typical failure for the smallest model: even when 
producing visible code, it frequently violates basic syntactic constraints and 
misinterprets the task specification.

\section{Prompts}

\subsection{Direct Instruction Prompts}

This is the Direct Instruction prompt setting on the IMDB dataset, where the task is to classify a given review as either positive or negative. Since the task is relatively simple, introducing the skill-based mechanism does not yield a significant performance improvement.

\begin{promptbox}{Direct Instruction Of IMDB}
Task: Classify the following movie review as positive or negative sentiment.

Review:
$<<<$Review Content$>>>$
\end{promptbox}

This is the Direct Instruction prompt setting for the FiNER dataset. The objective is to identify and classify XBRL tags. As shown, we provide not only the sentence but also the target numerical value appearing in that sentence. Because the \(139\) candidate tags correspond to specialized financial terminology, this task demands \textbf{strong} logical reasoning as well as substantial domain knowledge in finance.

\begin{promptbox}{Direct Instruction Of FiNER}
Given a sentence from financial documents, a target numeric entity from that sentence, and a list of candidate XBRL tags, choose the single best-matching XBRL tag.

Tag List (candidates): [InterestExpense, \ldots]

Inputs:

Sentence: $<<<$Sentence Content$>>>$

Target entity: $<<<$Numeric Entity$>>>$
\end{promptbox}

This is the Direct Instruction prompt setting for InsurBench. The task requires the agent to identify the key email(s) within a very long email thread and to make decisions on behalf of an insurance company. This setting evaluates the agent model's ability to localize salient information and to perform complex decision-making under long-context conditions.

\begin{promptbox}{Direct Instruction Of InsurBench}
Task: Given the full email thread in Email History, decide whether the insurance company must take action to reply.

Email History:
$<<<$Email History$>>>$
\end{promptbox}


\subsection{Skill Selection Prompts}

In this setting, we configure the model at initialization via the system prompt to prioritize skill selection rather than answering the user query directly. This system prompt is adapted from the agent skill system-agent implementation in the LangChain DeepAgent CLI, and we further refine it to better suit our use case.

\begin{promptbox}{Skill Selection System Prompt (Part 1)}
In order to complete the objective that the user asks of you, you have access to a number of skills.

\#\# Skills System

\textbf{Available Skills:}
{{Skill Context}}

\#\# Step-by-Step Process

\textbf{ALWAYS follow these exact steps:}
\end{promptbox}

\begin{promptbox}{Skill Selection System Prompt (Part 2)}
1. \textbf{Think step-by-step} about the user's request:
   - What is the main task?
   - Which skill domain does it match?

2. \textbf{Skill matching rules:}
   - \textbf{Multiple skills}: List names separated by commas (e.g., "langgraph-docs, sales-analytics")
   - \textbf{No skill matches}: Use empty list \texttt{[]}

3. \textbf{Generate response based on skills found:}

   | Skills Found | Message Content |
   |--------------|-----------------|
   | 1+ skills    | "Yes I need to read the skill information first because ..." |
   | No skills    | "I didn't find the right skill." |
\end{promptbox}

\begin{promptbox}{Skill Selection System Prompt (Part 3)}
\textbf{Final Output Format (Strict JSON)}

Your final response must \textbf{always} follow this JSON structure:

\{
  "Message": "Your complete response to the user query goes here.",
  "Skills": ["List of skills selected to complete the request, e.g., ['sales-analytics','sentiment-analytics']"]
\}

\#\# Examples
\end{promptbox}

\begin{promptbox}{Skill Selection System Prompt (Part 4)}
\#\# Examples

\textbf{User:} "How do I use LangGraph StateGraph?" \\
\textbf{Skills:} ["langgraph-docs"] \\
\textbf{JSON:} \{"Message": "Yes I need to read the skill information first because I need details on StateGraph from LangGraph docs.", "Skills": ["langgraph-docs"]\}

\textbf{User:} "Write a poem about cats" \\ 
\textbf{Skills:} [] \\
\textbf{JSON:} \{"Message": "I didn't find the right skill.", "Skills": []\}

\textbf{User:} "Write a report about LangGraph usage in sales." \\
\textbf{Skills:} ["langgraph-docs", "sales-analytics"] \\
\textbf{JSON:} \{"Message": "Yes I need to read the skill information first because it involves LangGraph documentation and sales analytics.", "Skills": ["langgraph-docs", "sales-analytics"]\}
\end{promptbox}


\begin{promptbox}{Skill Execution System Prompt (Part 1)}
In order to complete the objective that the user asks of you, you have access to a number of skills description.

\#\# Skills System
You have access to a skills library that provides specialized capabilities and domain knowledge.

\textbf{How to Use Skills:}

1. \textbf{Read the skill's full instructions}
2. \textbf{Follow the skill's instructions}: contains step-by-step workflows, best practices, and examples

\textbf{When to Use Skills:}
- User's request matches a skill's domain (e.g., "research X" $\rightarrow$ web-research skill)
- A skill provides proven patterns for complex tasks
\end{promptbox}

\begin{promptbox}{Skill Execution System Prompt (Part 2)}
\textbf{Example Workflow:}

User: "Can you research the latest developments in quantum computing?"

1. Check available skills description
2. Read the skill
3. Follow the skill's research workflow (search $\rightarrow$ organize $\rightarrow$ synthesize)
4. Make the final decision.

\textbf{Skill Information Collected}

{{Skill Context}}

Remember: Skills make you more capable and consistent. When in doubt, check if a skill exists for the task!

\#\# Output instructions

\textbf{Final Output Format (Strict JSON)}

\{
  "Message": Your message here.
\}
\end{promptbox}

\end{document}